\documentclass[10pt,twocolumn,letterpaper]{article}

\PassOptionsToPackage{table,xcdraw}{xcolor}

\usepackage{cvpr}              
\usepackage{booktabs}
\usepackage{geometry}
\usepackage{array}
\usepackage{multirow}
\usepackage{caption}
\usepackage{graphicx}
\usepackage{svg}
\usepackage{float}

\definecolor{cvprblue}{rgb}{0.21,0.49,0.74}
\usepackage[pagebackref,breaklinks,colorlinks,citecolor=cvprblue]{hyperref}

\def\paperID{*****} 
\def\confName{CVPR}
\def\confYear{2024}

\title{Analyzing CLIP's Performance Limitations in Multi-Object Scenarios: A Controlled High-Resolution Study}

\author{Reza Abbasi, Ali Nazari, Aminreza Sefid, Mohammadali Banayeeanzade, \\
Mohammad Hossein Rohban, Mahdieh Soleymani Baghshah\\
Sharif University of Technology, Tehran, Iran\\
{\tt\small \{reza.abbasi, ali.nazari02, aminreza.sefid, a.banayeean, rohban, soleymani\}@sharif.edu}
}

\begin{document}
\maketitle

\begin{abstract}
    Contrastive Language-Image Pre-training (CLIP) models have demonstrated remarkable performance in zero-shot classification tasks, yet their efficacy in handling complex multi-object scenarios remains challenging. This study presents a comprehensive analysis of CLIP's performance limitations in multi-object contexts through controlled experiments. We introduce two custom datasets, SimCO and CompCO, to evaluate CLIP's image and text encoders in various multi-object configurations. Our findings reveal significant biases in both encoders: the image encoder favors larger objects, while the text encoder prioritizes objects mentioned first in descriptions. We hypothesize these biases originate from CLIP's training process and provide evidence through analyses of the COCO dataset and CLIP's training progression. Additionally, we extend our investigation to Stable Diffusion models, revealing that biases in the CLIP text encoder significantly impact text-to-image generation tasks. Our experiments demonstrate how these biases affect CLIP's performance in image-caption matching and generation tasks, particularly when manipulating object sizes and their order in captions. This work contributes valuable insights into CLIP's behavior in complex visual environments and highlights areas for improvement in future vision-language models.
\end{abstract}    

\section{Introduction}
\label{sec:intro}
Contrastive Language-Image Pre-training (CLIP), introduced by OpenAI in 2021 \cite{radford2021learningtransferablevisualmodels}, represents a significant advancement in Vision-Language Models (VLMs). This innovative approach has demonstrated exceptional performance in zero-shot classification tasks, setting a new benchmark for the integration of visual and linguistic data \cite{Cherti_2023,gadre2023datacompsearchgenerationmultimodal,schuhmann2021laion400mopendatasetclipfiltered,thrush2022winoground}. However, its efficacy in processing complex multi-object scenarios remains an active area of research and development \cite{chen2024multi,ye2024beaf,kil2024compbench,castro2024clove,zarei2024understanding,tong2024eyes}.

While CLIP's capabilities are impressive, recent studies have identified notable limitations in its handling of scenarios involving images with multiple objects and their corresponding captions
\cite{Cherti_2023,yuksekgonul2023visionlanguagemodelsbehavelike,ma2023crepe}. These investigations have consistently highlighted multi-object scenarios as a particular challenge for the model. However, efforts to address these shortcomings have often proceeded without comprehensive evaluation and detailed analysis of the underlying causes, thereby necessitating a more thorough investigation to uncover the root causes and potential avenues for improving CLIP's performance in this critical domain.

\begin{figure} [t]
    \centering
    \includegraphics[width=\columnwidth]{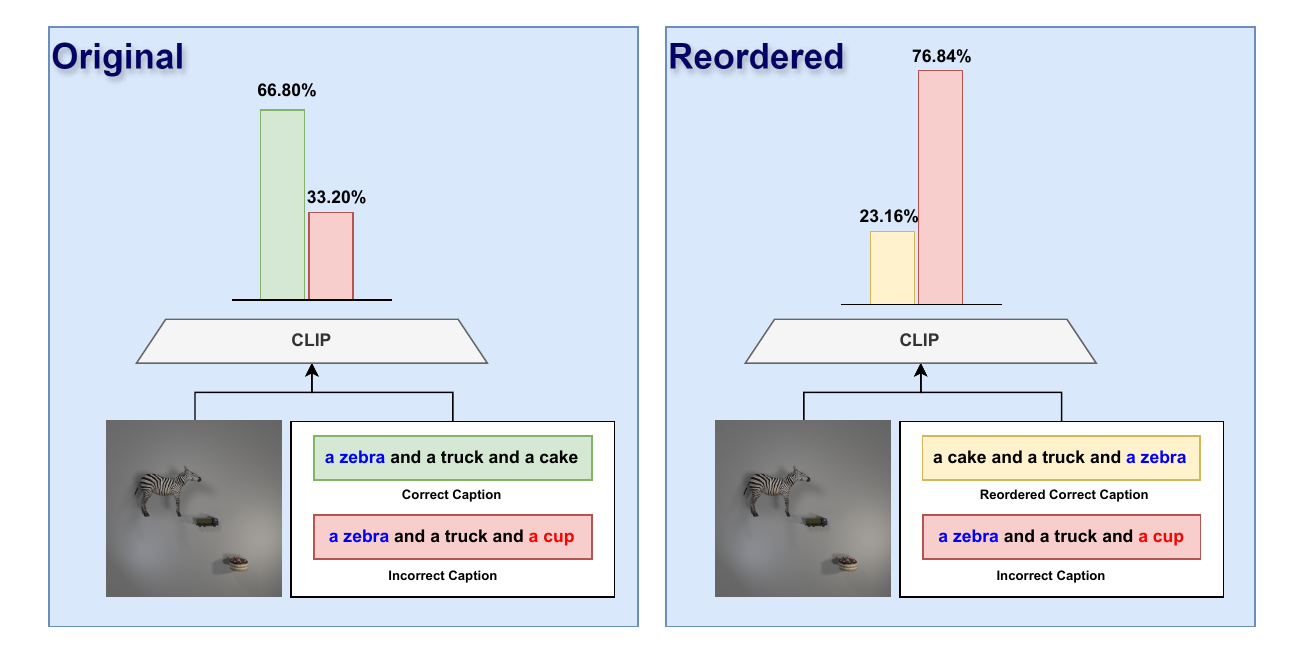}
    \caption{CLIP performance on multi-object image-caption matching. Left: Correct vs. incorrect captions with large object first. Right: Incorrect vs. reordered correct captions (large object last). Results show CLIP's bias for captions starting with larger objects, reducing accuracy when this order is altered.}
    \label{fig:first_fig}
\end{figure}

In our work, we tried to fill this gap by conducting a comprehensive analysis of CLIP's performance in multi-object scenarios through a series of controlled fine-grained studies. Our research focuses on several key aspects:

\begin{enumerate}
    \item Evaluating CLIP's performance using two synthetic datasets, SimCO and CompCO, designed specifically for controlled multi-object scenarios.
    \item Analyzing both CLIP's image encoder and text encoder biases when processing multi-object scenes and descriptions.
    \item Examining CLIP's biases in multi-object processing, their origins, and implications for image-caption matching and text-to-image generation tasks.
\end{enumerate}

\section{Methodology}
\label{sec:methodology}
\subsection{Dataset Design}
\begin{figure}[t]
    \centering
    \includegraphics[width=\columnwidth]{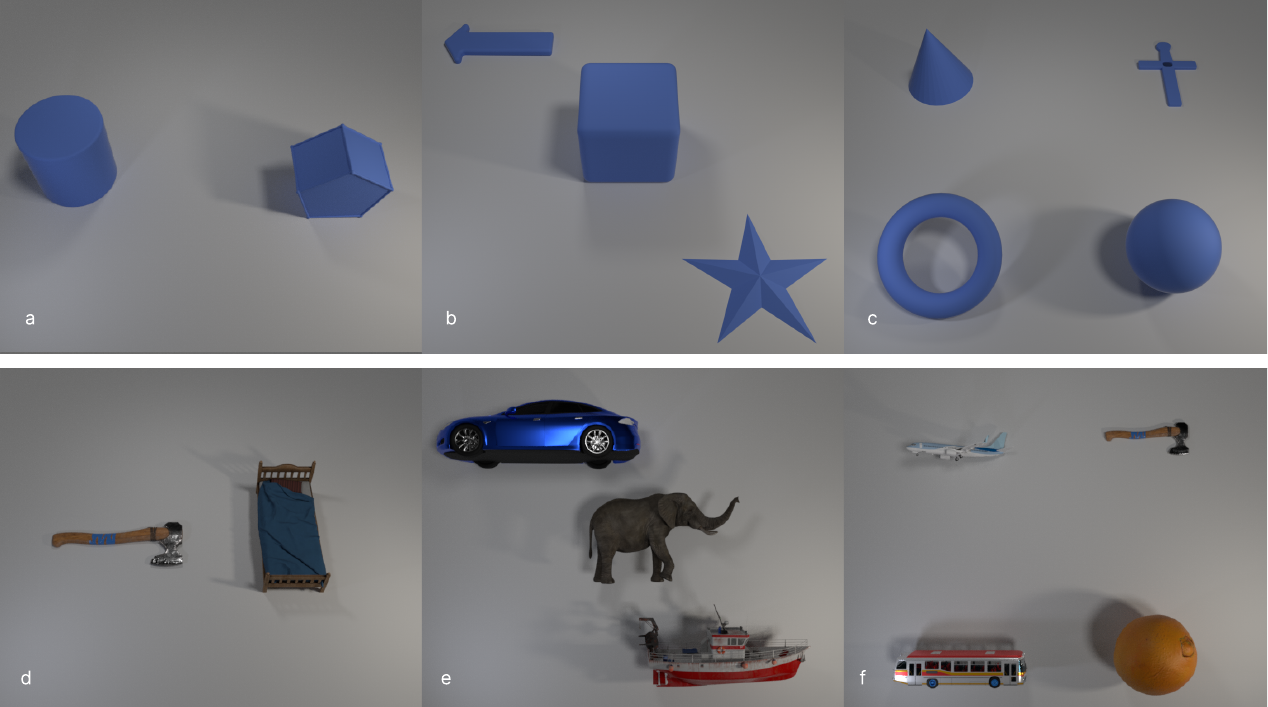}
    \caption{Example images from the SimCO and CompCO datasets.}
    \label{fig:datasets}
\end{figure}

To evaluate CLIP models in multi-object scenarios under controlled conditions, we created two datasets: \textbf{SimCO} and \textbf{CompCO}, using Blender for precise control over objects' number, location, and size (see Fig \ref{fig:datasets}, \ref{fig:simco}, \ref{fig:comco}).

\begin{itemize}
    \item \textbf{SimCO Dataset}: Inspired by the CLEVER dataset \cite{johnson2016clevrdiagnosticdatasetcompositional}, SimCO includes 17 basic geometric objects, compared to CLEVER, which only contains 3 such objects. This dataset tests the model’s ability to handle simple shapes and configurations under controlled conditions.
    
    \item \textbf{CompCO Dataset}: Derived from the COCO dataset \cite{lin2015microsoftcococommonobjects}, CompCO comprises 72 complex and commonly occurring objects. It evaluates the model's performance with more realistic and intricate object arrangements, better matching images in the real-world scenarios.
\end{itemize}

Both datasets contain images with 2, 3, 4, and 5 objects, each paired with a caption accurately describing the objects. This ensures high control and minimizes confounding factors, providing a robust platform for evaluating CLIP models. Additional information can be found in the \ref{app:dataset}.

\subsection{Experimental Setup}

To evaluate CLIP's performance in multi-object scenarios, we designed a series of experiments focusing on both the image and text encoders. Our goal is to assess how each encoder processes and represents multiple objects in their respective modalities.

\subsubsection{Image Encoder Analysis}
To investigate potential biases related to object size, we conducted two experiments:

\begin{enumerate}
    \item \textbf{Image-based Object Classification (IOC):} \textbf{\textit{For each object}} in a multi-object image, we trained a single-layer classifier on the vector representations generated by the CLIP image encoder. We then calculated the classification accuracy on test set for individual objects within the images.
    
    \item \textbf{Image-based Object Retrieval (IOR):} For each multi-object image, we identified the most similar single-object images using the CLIP image encoder's representations. We then computed the percentage of cases where the nearest single-object image corresponded to each object in the multi-object image, categorized by size (e.g., largest object, smaller object). 
    This allowed us to quantify CLIP's tendency to prioritize certain objects based on their relative sizes within the multi-object scene.
\end{enumerate}


\subsubsection{Text Encoder Analysis}

To assess how object mention order affects text representations, we performed two experiments:

\begin{enumerate}
    \item \textbf{Text-based Object Classification (TOC):} \textbf{\textit{For each object}} mentioned in a multi-object text description, we train a single-layer binary classifier on the vector representations produced by the CLIP text encoder. We then assess the classification accuracy for individual objects within the text.
    
    \item \textbf{Text-based Object Retrieval (TOR):} For each multi-object text description, we identify the most similar single-object text using the CLIP text encoder's representation. We then calculate the percentage of cases where the nearest single-object text matches to each mentioned object in the multi-object text description (e.g., first mentioned, second mentioned, etc.).
\end{enumerate}

\section{Results and Analysis}
\label{sec:results}
Our experiments revealed significant biases in both the text and image encoders of the CLIP model. This section presents our findings, organized by the encoder type and experiment.

\subsection{Text Encoder Analysis}

\subsubsection{Text-based Object Classification (TOC)}

The performance of the text encoder in the TOC experiment demonstrated a significant bias towards the first object mentioned in the text descriptions. As shown in Table \ref{tab:tor_results}, the classification accuracy for the first object was considerably higher than for subsequent objects. This suggests that the first object mentioned in a textual description is represented more prominently in the final textual representation.

\subsubsection{Text-based Object Retrieval (TOR)}

The TOR experiment further reinforced the presence of bias in the text encoder. Table \ref{tab:tor_results} presents the retrieval accuracy, where the first object mentioned in the descriptions consistently showed the highest retrieval accuracy. This indicates that the initial object exerts a dominant influence on the overall text representation, making it more likely to be retrieved accurately compared to subsequent objects.

\begin{table}[ht]
\centering
\scriptsize
\setlength{\tabcolsep}{3pt}
\renewcommand{\arraystretch}{1.2}
\caption{Performance of various CLIP models on TOC and TOR for ComCO datasets}
\label{tab:tor_results}
\begin{tabular}{llcccc}
\toprule
\rowcolor[HTML]{EFEFEF}
Task & Model & \textbf{First Obj} & \textbf{Second Obj} & \textbf{Third Obj} & \textbf{Fourth Obj} \\ 
\midrule
\multirow{5}{*}{TOC} 
 & \textit{CLIP openAI} & \textbf{87.17} & 30.60 & 31.69 & 74.49 \\
 & \textit{CLIP LAION}\cite{schuhmann2022laion} & \textbf{98.89} & 31.64 &20.90 & 47.76 \\
 & \textit{CLIP Datacomp}\cite{gadre2024datacomp} & \textbf{99.46} & 22.82 & 32.93 & 58.18 \\
 & \textit{SIGLIP}\cite{zhai2023sigmoid} & \textbf{97.27} & 72.51 & 33.25 & 5.79 \\
 & \textit{NegCLIP}\cite{yuksekgonul2023and} & \textbf{98.73} & 28.05 & 30.83 & 43.82 \\
\midrule
\multirow{5}{*}{TOR}   
 & \textit{CLIP openAI} & \textbf{48.20} & 26.01 & 10.74 & 8.74 \\
 & \textit{CLIP LAION} & \textbf{63.96} & 21.59 & 10.68 & 3.76 \\
 & \textit{CLIP Datacomp} & \textbf{71.13} & 16.26 & 8.74 & 3.87 \\
 & \textit{SIGLIP} & \textbf{58.11} & 21.16 & 10.99 & 9.73 \\
 & \textit{NegCLIP} & \textbf{51.63} & 28.92 & 14.86 & 4.59 \\
\bottomrule
\end{tabular}
\end{table}

\subsection{Image Encoder Analysis}

\subsubsection{Image-based Object Classification (IOC)}
IOC experiment revealed that larger objects in an image significantly influence the final visual representation more than smaller objects. This size-related bias is evident across various models and datasets. As detailed in Table \ref{tab:ior_size}, the classification accuracy for larger objects was consistently higher, indicating that the image encoder prioritizes these objects in its representations. 

\subsubsection{Image-based Object Retrieval (IOR)}
The IOR experiment confirmed the significant influence of larger objects on the image encoder's performance. Table \ref{tab:ior_size} shows that larger objects in multi-object images were more frequently and accurately identified in single-object image searches. This suggests a strong size-related bias in the image encoder, where larger objects are given more weight in the final image representation.  



\begin{table}[ht]
\centering
\scriptsize
\setlength{\tabcolsep}{3pt}
\renewcommand{\arraystretch}{1.2}
\caption{Performance of various CLIP models on IOC and IOR for ComCO dataset. The second object is the largest.}
\label{tab:ior_size}
\begin{tabular}{llcccc}
\toprule
\rowcolor[HTML]{E4E8F2}
Task & Model & \textbf{Large Object} & \textbf{Small Obj 1} & \textbf{Small Obj 2} & \textbf{Small Obj 3} \\ 
\midrule
\multirow{5}{*}{IOC} 
 & \textit{CLIP openAI} & \textbf{99.94} & 19.32 & 21.89 & 22.39 \\
 & \textit{CLIP LAION} & \textbf{100.0} & 19.76 & 17.57 & 18.89 \\
 & \textit{CLIP Datacomp} & \textbf{100.0} & 20.64 & 21.01 & 19.01 \\
 & \textit{SIGLIP} & \textbf{100.0} & 18.95 & 15.57 & 17.57 \\
 & \textit{NegCLIP} & \textbf{100.0} & 21.89 & 23.64 & 31.33 \\
\midrule
\multirow{5}{*}{IOR}   
 & \textit{CLIP openAI} & \textbf{67.86} & 14.29 & 7.14 & 10.71 \\
 & \textit{CLIP LAION} & \textbf{91.78} & 5.48 & 2.74 & 0.00 \\
 & \textit{CLIP Datacomp} & \textbf{93.30} & 3.91 & 1.12 & 1.68 \\
 & \textit{SIGLIP} & \textbf{94.03} & 2.24 & 1.49 & 2.24 \\
 & \textit{NegCLIP} & \textbf{79.55} & 0.00 & 2.27 & 18.19 \\
\bottomrule
\end{tabular}
\end{table}

Experiments on SimCO and additional experiments can be found in \ref{app:toc},\ref{app:tor},\ref{app:ioc},\ref{app:ior},\ref{app:toc-long},\ref{app:tor-long}.

\section{Bias Origin Hypothesis}

This section explores the potential origins of these biases and presents evidence supporting our hypotheses.

\subsection{Image Encoder Bias}
The image domain's bias towards larger objects can be attributed to the structure of vision transformers (ViT) in CLIP's image encoder. Larger objects occupy more patches in the ViT's image division, potentially exerting greater influence on the final CLS token representation.

\subsection{Text Encoder Bias}
The bias in the text domain, favoring the first-mentioned object, is more subtle and warrants deeper examination. We hypothesize that this bias originates from CLIP's contrastive training process, which may transfer the image-side bias (preference for larger objects) to the text side. This hypothesis is founded on two key observations:
\begin{enumerate}
\item In CLIP training datasets, larger objects tend to be mentioned earlier in text descriptions.
\item The contrastive training procedure aligns text and image embeddings, potentially facilitating the transfer of biases between modalities.
\end{enumerate}
To validate this hypothesis, we conducted two experimental studies:


\subsection{Experiment 1: COCO Dataset Analysis}
\begin{figure} [h!]
    \centering
    \includegraphics[width=\columnwidth]{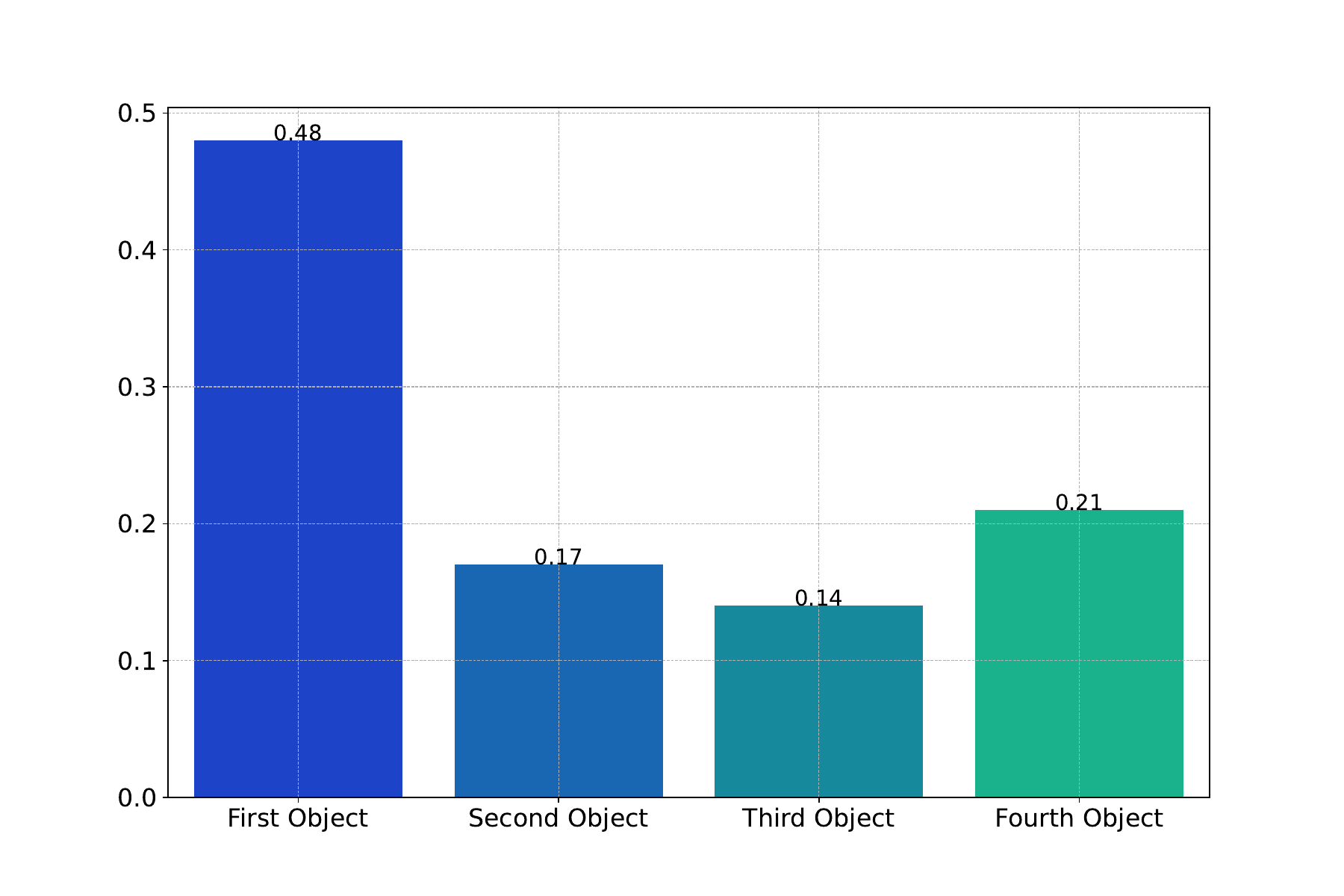}
    \caption{In the COCO dataset, the larger objects in an image are typically mentioned earlier in the captions}
    \label{fig:coco_analysis}
\end{figure}

We analyzed the COCO dataset, which is similar to CLIP training datasets, to investigate object size and text position correlation. Our method used LLaMA3 \cite{touvron2023llama} to extract objects from captions and OWL2 \cite{minderer2024scaling} for object detection in images. We calculated bounding box areas and examined the correlation between object size and text position. Figure \ref{fig:coco_analysis} shows the percentage of cases where larger objects appeared earlier in the text. More information can be found in the \ref{app:coco-anlysis} .

 \vspace{-0.1in}
\subsection{Experiment 2: CLIP Training Progression}
\label{sec:bias_origin}
We examined text-side bias development during CLIP training using TOR analysis at five training stages (2, 4, 6, 8, and 10 billion samples) on the LAION dataset.

\begin{figure}[h!]
    \centering
    \includegraphics[width=\columnwidth]{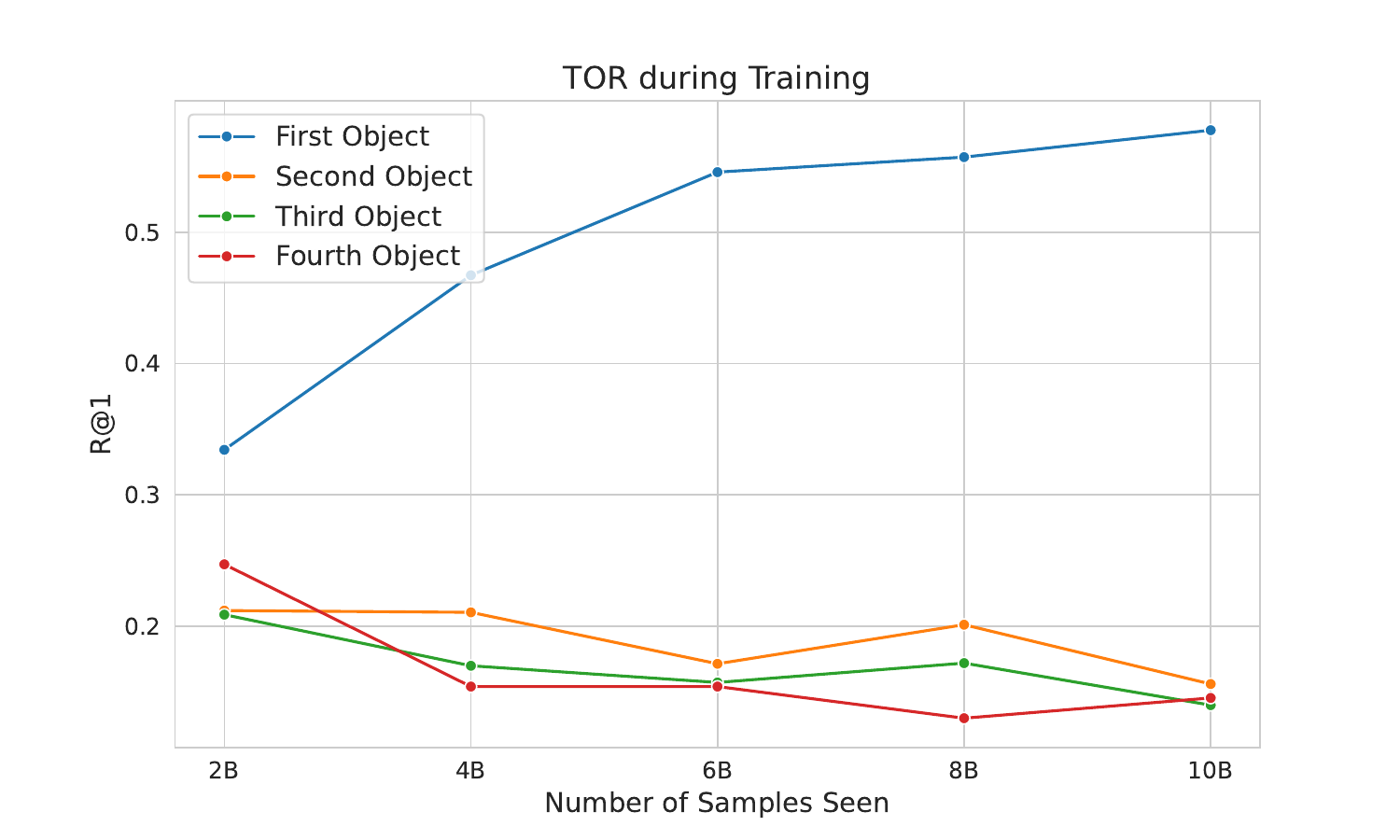}
    \caption{Evolution of TOR rate across training stages}
    \label{fig:training_progression}
\end{figure}

Figure \ref{fig:training_progression} shows the evolution of text-side bias through TOR rates for different objects. The graph reveals a steady increase in TOR rate for the first object as training progresses.

These results suggest that the bias towards the first object in text representations developed gradually during training, supporting our hypothesis that text-side bias is induced during contrastive learning, likely aligning with the image encoder's bias towards larger, more prominent objects.

\section{Practical Impacts of Encoder Biases}
\label{sec:impact}

\subsection{Image-Text Matching}
In the previous section, we showed that biases exist in both the image and text encoders of the CLIP model. Here, we illustrate how these biases affect the model's performance in the image-caption matching task, leading to a noticeable performance decrease.

We conducted experiments with images containing 2 to 5 objects. Each image had two captions: one correct and one incorrect. The correct caption listed all objects, while the incorrect one replaced one object with a different one. We tested two scenarios:

\begin{enumerate}
    \item \textbf{Original:} The object order in the captions is the same up to the last object, with the larger object first. The last object differs between captions.
    \item \textbf{Reordered:} The order in the correct caption is altered by moving the largest object to the end (see Figure \ref{fig:first_fig} for an example).
\end{enumerate}

As shown in Table \ref{table:performance_drop}, model performance significantly drops when the incorrect caption places the large object first. Detailed results for other scenarios and dataset are provided in the \ref{app:imtexmatch}.

These findings highlight how biases in the text and image encoders lead to a substantial performance decrease in multi-object scenarios.

\begin{table}[ht]
\centering
\scriptsize
\setlength{\tabcolsep}{6pt}
\renewcommand{\arraystretch}{1.2}
\caption{Performance of Various Models on SimCO and ComCO Datasets}
 \label{table:performance_drop}
\begin{tabular}{llcc}
\toprule
\rowcolor[HTML]{EFEFEF}
Dataset & Model & \textbf{Original} & \textbf{Reordered} \\ 
\midrule
\multirow{5}{*}{ComCO}   
& \textit{CLIP openAI} & 80.48 & 60.71 \\
& \textit{CLIP LAION} & 82.82 & 50.62 \\
& \textit{CLIP Datacomp} & 86.24 & 63.99 \\
 & \textit{SIGLIP} & 84.73 & 74.09 \\
 & \textit{NegCLIP} & 76.68 & 46.94 \\
\bottomrule
\end{tabular}
\end{table}

\subsection{Text to image geneation}
To observe CLIP's bias impact on Stable Diffusion models, we generated 1,000 multi-object images using prompts with four objects from the COCO dataset. LLAVA model \cite{liu2024visual} evaluated object presence in generated images. Results in Table \ref{tab:tor_results2} show the bias's clear impact, with earlier-mentioned objects appearing more frequently.

\begin{table}[ht]
\centering
\scriptsize
\setlength{\tabcolsep}{3pt}
\renewcommand{\arraystretch}{1.2}
\caption{Object presence in Stable Diffusion-generated images }
\label{tab:tor_results2}
\begin{tabular}{lcccc}
\toprule
\rowcolor[HTML]{EFEFEF}
Model & \textbf{First Obj} & \textbf{Second Obj} & \textbf{Third Obj} & \textbf{Fourth Obj} \\ 
\midrule 
\textit{SD v1.4} & 17.8 & 14.6 & 13.4 & 12.0 \\
\textit{SD V2} & 18.7 & 16.3 & 14.3 & 14.3 \\
\textit{SD-XL} \cite{podell2023sdxl} & 33.3 & 29.5 & 25.6 & 25.5 \\
\bottomrule
\end{tabular}
\end{table}

While multi-object image generation remains challenging, these results demonstrate the text encoder bias's impact on Stable Diffusion models.   
\section{Conclusion}
Our study reveals significant biases in CLIP's image and text encoders, favoring larger objects and first-mentioned items respectively. These biases, demonstrated through our SimCO and CompCO datasets, substantially impact CLIP's performance in multi-object scenarios. The observed performance drops when manipulating object sizes and mention order underscore CLIP's limitations in handling complex visual environments. These findings highlight the need for more balanced training approaches in vision-language models to mitigate such biases. Future work should focus on developing techniques to address these limitations, advancing the field towards more robust and versatile AI systems capable of accurately interpreting multi-faceted real-world information.

{
    \small
    \bibliographystyle{ieeenat_fullname}
    \bibliography{main}
}
\clearpage
\setcounter{page}{1}
\maketitlesupplementary

\section{The SIMCO and ComCO Datasets}
\label{app:dataset}
\subsection{The SIMCO Dataset}
The SIMCO dataset comprises 17 objects. These 17 objects are:

\begin{center}
\begin{tabular}{lll}
Cube & Sphere & Cylinder \\
Mug & Pentagon & Heart \\
Cone & Pyramid & Diamond \\
Moon & Cross & Snowflake \\
Leaf & Arrow & Star \\
Torus & Pot &
\end{tabular}
\end{center}

\begin{figure*}[htbp]
    \centering
    \includegraphics[width=0.9\linewidth]{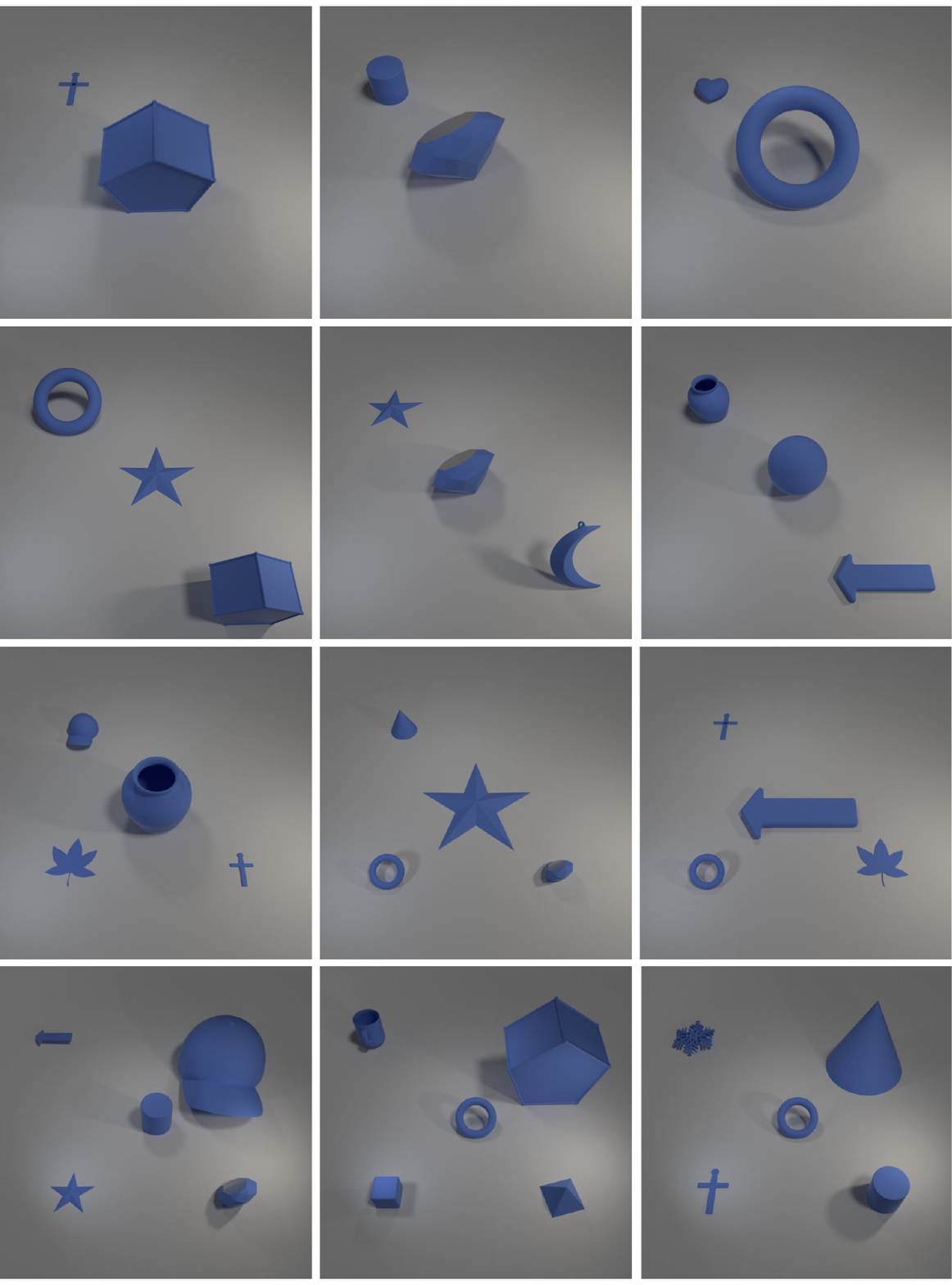}
    \caption{Examples of SimCO dataset}
    \label{fig:simco}
\end{figure*}

Using Blender software, a collection of images containing 2 to 5 objects has been created from these 17 objects. The total number of images in this dataset is approximately 85,000. Examples of these images can be seen in Figure \ref{fig:simco}.

\subsection{The ComCO Dataset}

The ComCO dataset contains 72 objects, as listed below:
\begin{center}
\scriptsize 
\begin{tabular}{@{} l @{\hspace{2em}} l @{\hspace{2em}} l @{\hspace{2em}} l @{}}
person & bicycle & car & motorcycle \\
airplane & bus & train & truck \\
boat & traffic light & fire hydrant & street sign \\
stop sign & parking meter & bench & bird \\
cat & dog & horse & sheep \\
cow & dining table & cell phone & elephant \\
bear & zebra & giraffe & hat \\
backpack & umbrella & shoe & eye glasses \\
handbag & tie & suitcase & frisbee \\
skis & snowboard & kite & baseball bat \\
baseball glove & tennis racket & wine glass & hot dog \\
potted plant & teddy bear & hair drier & hair brush \\
skateboard & surfboard & bottle & plate \\
cup & fork & knife & spoon \\
bowl & banana & apple & sandwich \\
orange & broccoli & carrot & pizza \\
donut & cake & chair & couch \\
bed & mirror & window & desk \\
toilet & door & tv & laptop \\
mouse & remote & keyboard & microwave \\
oven & toaster & sink & refrigerator \\
blender & book & clock & vase \\
scissors & toothbrush & & \\
\end{tabular}
\end{center}

\begin{figure*}[htbp]
    \centering
    \includegraphics[width=0.9\linewidth]{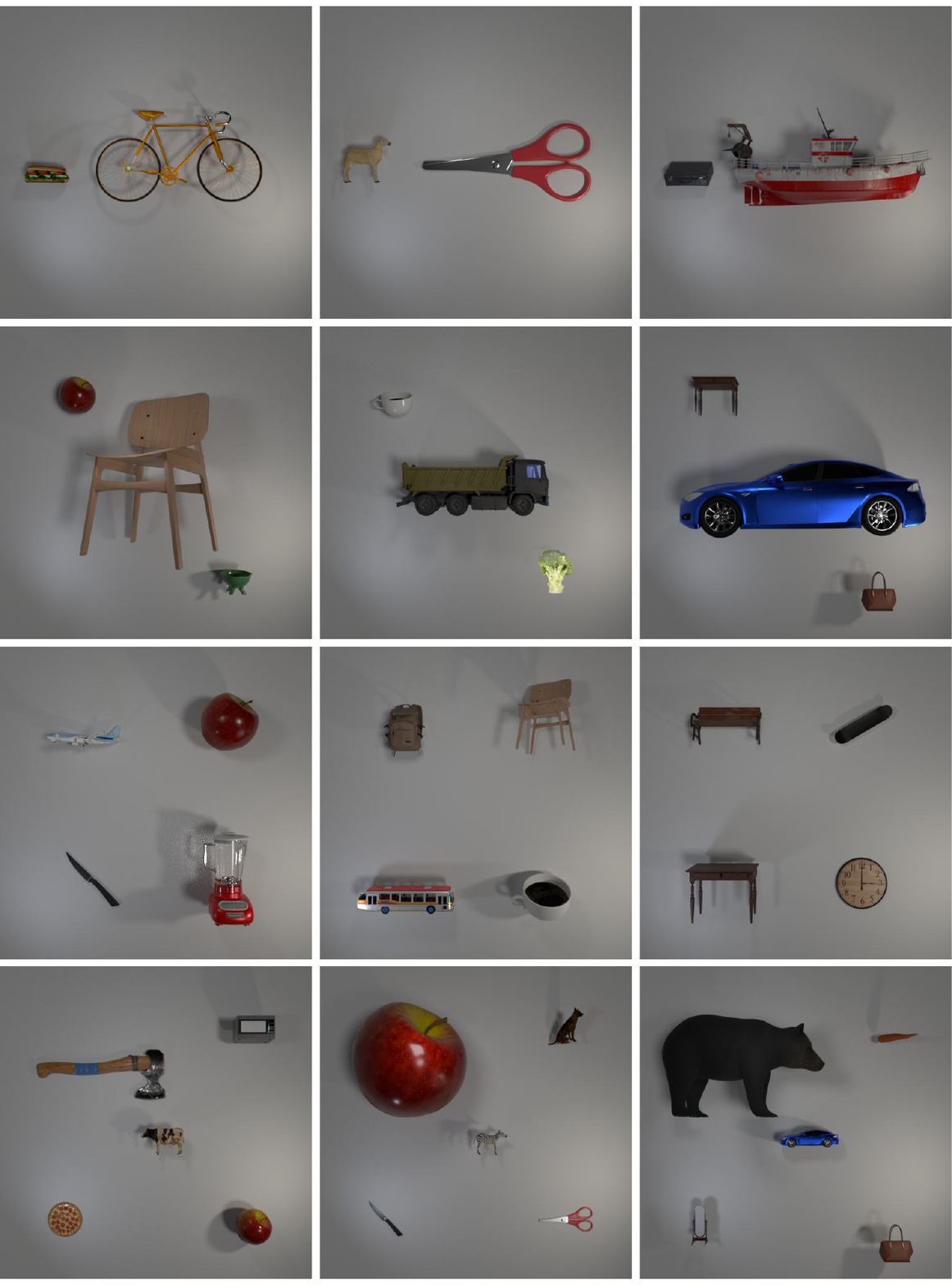}
    \caption{Examples of ComCO dataset}
    \label{fig:comco}
\end{figure*}

In this dataset, a collection of images containing 2 to 5 different objects has also been generated. The total number of images in this dataset is approximately 190,000. Various examples from this dataset can be seen in Figure \ref{fig:comco}.

\section{Text-based Object Classification}
\label{app:toc}
We conducted the TOC experiment on various models under different scenarios, and the results are presented in Table \ref{tab:toc_total}. This experiment was repeated on both the SIMCO and ComCO datasets.

\begin{table*}[htbp]
    \centering
    \scriptsize
    \setlength{\tabcolsep}{4pt}
    \renewcommand{\arraystretch}{1.1}
    \caption{Text-based Object Classification}
    \label{tab:toc_total}
    \begin{tabular}{lllccccc}
    \toprule
    \rowcolor[HTML]{E4E8F2}
    Number of Objects & Dataset & Model & \textbf{First Object} & \textbf{Second Object} & \textbf{Third Object} & \textbf{Fourth Object} & \textbf{Fifth Object} \\ 
    \midrule
    \multirow{16}{*}{n = 2}  & \multirow{8}{*}{SimCO}    
    &  \textit{ViT-H-14 (DFN)}& 99.86 & 97.09 & - & - & - \\
    & & \textit{ViT-SO400M-SigLIP} & 98.67 & 91.29 & - & - & - \\
    & & \textit{ViT-L-14 (datacomp)}& 99.76 & 96.77 & - & - & - \\
    & & \textit{xlm-roberta-large-ViT-H-14}& 99.03 & 89.87 & - & - & - \\
    & & \textit{ViT-L-14 (laion2b)} & 99.70 & 97.57 & - & - & - \\
    & & \textit{ViT-L-14 (openai)} & 97.62 & 91.30 & - & - & - \\
    & & \textit{ViT-B-32 (openai)} & 96.85 & 73.00 & - & - & - \\
    & & \textit{NegCLIP} & 98.19 & 84.43  & - & - & - \\

    \cmidrule{2-8}
    
    & \multirow{8}{*}{ComCO}    
    &  \textit{ViT-H-14 (DFN)}& 99.90 & 96.56 & - & - & - \\
    & & \textit{ViT-SO400M-SigLIP} & 98.47 & 93.18 & - & - & - \\
    & & \textit{ViT-L-14 (datacomp)}& 99.74 & 96.86 & - & - & - \\
    & & \textit{xlm-roberta-large-ViT-H-14}& 99.16 & 91.57 & - & - & - \\
    & & \textit{ViT-L-14 (laion2b)} & 99.72 & 96.24 & - & - & - \\
    & & \textit{ViT-L-14 (openai)} & 97.93 & 96.69 & - & - & - \\
    & & \textit{ViT-B-32 (openai)} & 96.86 & 85.42 & - & - & - \\
    & & \textit{NegCLIP} & 99.30 & 92.09 & - & - & - \\
    
    \midrule 
    \multirow{16}{*}{n = 3} & \multirow{8}{*}{SimCO}   
    &  \textit{ViT-H-14 (DFN)}& 99.46 & 60.47 & 76.99 & - & - \\
    & & \textit{ViT-SO400M-SigLIP} & 98.23 & 71.42 & 45.80 & - & - \\
    & & \textit{ViT-L-14 (datacomp)}& 99.49 & 45.80 & 78.66 & - & - \\
    & & \textit{xlm-roberta-large-ViT-H-14}& 99.26 & 49.08 & 64.07 & - & - \\
    & & \textit{ViT-L-14 (laion2b)} & 98.93 & 56.87 & 72.37 & - & - \\
    & & \textit{ViT-L-14 (openai)} & 91.87 & 50.75 & 68.38 & - & - \\
    & & \textit{ViT-B-32 (openai)} & 92.55 & 38.61 & 52.94 & - & - \\
    & & \textit{NegCLIP} & 95.80 & 44.70 & 59.11 & - & - \\

    \cmidrule{2-8}
    & \multirow{8}{*}{ComCO}    
    &  \textit{ViT-H-14 (DFN)}& 99.73 & 59.80 & 73.63 & - & - \\
    & & \textit{ViT-SO400M-SigLIP} & 96.94 & 70.26 & 29.28 & - & - \\
    & & \textit{ViT-L-14 (datacomp)}& 99.53 & 45.13 & 74.15 & - & - \\
    & & \textit{xlm-roberta-large-ViT-H-14}& 99.20 & 53.34 & 57.15 & - & - \\
    & & \textit{ViT-L-14 (laion2b)} & 99.26 & 58.58 & 64.74 & - & - \\
    & & \textit{ViT-L-14 (openai)} & 90.86 & 49.67 & 83.49 & - & - \\
    & & \textit{ViT-B-32 (openai)} & 87.97 & 45.77 & 63.13 & - & - \\
    & & \textit{NegCLIP} & 56.94 & 98.03 & 56.66 & - & - \\
     
    \midrule
    \multirow{16}{*}{n = 4} & \multirow{8}{*}{SimCO}   
    
    &  \textit{ViT-H-14 (DFN)}& 99.46 & 34.57 & 36.73 & 62.35 & - \\
    & & \textit{ViT-SO400M-SigLIP} & 98.23 & 69.91 & 26.10 & 6.54 & - \\
    & & \textit{ViT-L-14 (datacomp)}& 99.00 & 23.76 & 35.55 & 60.55 & - \\
    & & \textit{xlm-roberta-large-ViT-H-14}& 99.26 & 27.97 & 28.84 & 48.34 & - \\
    & & \textit{ViT-L-14 (laion2b)} & 98.82 & 34.21 & 31.41 & 54.73 & - \\
    & & \textit{ViT-L-14 (openai)} & 90.48 & 35.19 & 30.50 & 59.29 & - \\
    & & \textit{ViT-B-32 (openai)} & 90.76 & 22.77 & 25.36 & 40.45 & - \\
    & & \textit{NegCLIP} & 96.50 & 9.33 & 4.79 & 15.58 & - \\

    \cmidrule{2-8}
    & \multirow{8}{*}{ComCO}    
    &  \textit{ViT-H-14 (DFN)}& 99.76 & 31.74 & 35.29 & 54.82 & - \\
    & & \textit{ViT-SO400M-SigLIP} & 97.27 & 72.51 & 33.25 & 5.79 & - \\
    & & \textit{ViT-L-14 (datacomp)}& 99.46 & 22.82 & 32.93 & 58.18 & - \\
    & & \textit{xlm-roberta-large-ViT-H-14}& 99.60 & 26.27 & 26.20 & 36.51 & - \\
    & & \textit{ViT-L-14 (laion2b)} & 98.89 & 31.64 & 20.90 & 47.76 & - \\
    & & \textit{ViT-L-14 (openai)} & 87.17 & 30.60 & 31.69 & 74.49 & - \\
    & & \textit{ViT-B-32 (openai)} & 88.24 & 24.23 & 28.30 & 49.82 & - \\
    & & \textit{NegCLIP} & 98.73 & 28.05 & 30.83 & 43.82 & - \\
     
    \midrule
    \multirow{16}{*}{n = 5}& \multirow{8}{*}{SimCO}

    &  \textit{ViT-H-14 (DFN)}& 99.00 & 24.30 & 22.33 & 27.23 & 53.03 \\
    & & \textit{ViT-SO400M-SigLIP} & 97.79 & 71.67 & 27.41 & 6.29 & 6.48 \\
    & & \textit{ViT-L-14 (datacomp)}& 98.89 & 16.51 & 21.29 & 26.92 & 48.52 \\
    & & \textit{xlm-roberta-large-ViT-H-14}& 99.46 & 17.15 & 16.63 & 20.18 & 35.64 \\
    & & \textit{ViT-L-14 (laion2b)} & 98.43 & 25.51 & 19.81 & 23.15 & 41.07 \\
    & & \textit{ViT-L-14 (openai)} & 89.79 & 26.33 & 20.74 & 24.69 & 50.29 \\
    & & \textit{ViT-B-32 (openai)} & 92.73 & 15.67 & 17.03 & 19.58 & 33.62 \\
    & & \textit{NegCLIP} & 96.83 & 15.50 & 17.54 & 22.58 & 36.40 \\

    \cmidrule{2-8}
    & \multirow{8}{*}{ComCO}    
    &  \textit{ViT-H-14 (DFN)}& 99.80 & 19.44 & 20.79 & 24.86 & 42.38 \\
    & & \textit{ViT-SO400M-SigLIP} & 97.63 & 70.57 & 32.34 & 5.42 & 5.72 \\
    & & \textit{ViT-L-14 (datacomp)}& 99.13 & 14.75 & 19.89 & 25.72 & 47.11 \\
    & & \textit{xlm-roberta-large-ViT-H-14}& 99.40 & 18.21 & 15.47 & 18.05 & 26.12 \\
    & & \textit{ViT-L-14 (laion2b)} & 98.76 & 20.91 & 18.11 & 20.77 & 33.54 \\
    & & \textit{ViT-L-14 (openai)} & 86.13 & 22.11 & 19.43 & 28.03 & 68.37 \\
    & & \textit{ViT-B-32 (openai)} & 91.20 & 15.56 & 13.31 & 19.66 & 39.39 \\
    & & \textit{NegCLIP} & 99.03 & 16.69 & 16.51 & 22.26 & 34.29 \\

    \bottomrule
    \end{tabular}
\end{table*}

\section{Text-based Object Retrieval}
\label{app:tor}
We repeated the TOR experiment on various models across scenarios with captions containing 2 to 5 objects. This was done to confirm the presence of the discovered bias. The complete results of this experiment, which was conducted on both the SIMCO and ComCO datasets, can be observed in Table \ref{tab:tor_total}.

\begin{table*}[htbp]
    \centering
    \scriptsize
    \setlength{\tabcolsep}{3pt}
    \renewcommand{\arraystretch}{1.1}
    \caption{Text-based Object Retrieval}
    \label{tab:tor_total}
    \begin{tabular}{llcccccc}
    \toprule
    \rowcolor[HTML]{E4E8F2}
    Number of Objects & Dataset & Model & \textbf{First Object} & \textbf{Second Object} & \textbf{Third Object} & \textbf{Fourth Object} & \textbf{Fifth Object} \\ 
    \midrule
    \multirow{16}{*}{n = 2}  & \multirow{8}{*}{SimCO}    
    &  \textit{ViT-H-14 (DFN)}& 69.18 & 30.82 & - & - & - \\
    & & \textit{ViT-SO400M-SigLIP} & 68.87 & 31.13 & - & - & - \\
    & & \textit{ViT-L-14 (datacomp)}& 69.93 & 30.07 & - & - & - \\
    & & \textit{xlm-roberta-large-ViT-H-14}& 78.95 & 21.05 & - & - & - \\
    & & \textit{ViT-L-14 (laion2b)} & 68.66 & 31.34 & - & - & - \\
    & & \textit{ViT-L-14 (openai)} & 75.82 & 24.18 & - & - & - \\
    & & \textit{ViT-B-32 (openai)} & 81.05 & 18.95 & - & - & - \\
    & & \textit{NegCLIP} & 77.78 & 22.22 & - & - & - \\

    \cmidrule{2-8}
    
    & \multirow{8}{*}{ComCO}    
    &  \textit{ViT-H-14 (DFN)}& 70.87 & 29.13 & - & - & - \\
    & & \textit{ViT-SO400M-SigLIP} & 67.56 & 32.44 & - & - & - \\
    & & \textit{ViT-L-14 (datacomp)}& 70.37 & 26.93 & - & - & - \\
    & & \textit{xlm-roberta-large-ViT-H-14}& 59.15 & 40.85 & - & - & - \\
    & & \textit{ViT-L-14 (laion2b)} & 70.84 & 29.16 & - & - & - \\
    & & \textit{ViT-L-14 (openai)} & 66.03 & 33.97 & - & - & - \\
    & & \textit{ViT-B-32 (openai)} & 61.62  & 38.38 & - & - & - \\
    & & \textit{NegCLIP} & 64.13 & 35.87 & - & - & - \\
    
    \midrule 
    \multirow{16}{*}{n = 3} & \multirow{8}{*}{SimCO}   
    &  \textit{ViT-H-14 (DFN)}& 62.05 & 18.07 & 19.88 & - & - \\
    & & \textit{ViT-SO400M-SigLIP} & 58.05 & 20.50 & 21.46 & - & - \\
    & & \textit{ViT-L-14 (datacomp)}& 61.68 & 20.35 & 17.96 & - & - \\
    & & \textit{xlm-roberta-large-ViT-H-14}& 66.75 & 23.86 & 9.39 & - & - \\
    & & \textit{ViT-L-14 (laion2b)} & 62.31 & 12.56 & 25.13 & - & - \\
    & & \textit{ViT-L-14 (openai)} & 65.71 & 16.67 & 17.62 & - & - \\
    & & \textit{ViT-B-32 (openai)} & 74.23 & 13.62 & 12.15 & - & - \\
    & & \textit{NegCLIP} & 77.43 & 13.75 & 8.83 & - & - \\

    \cmidrule{2-8}
    & \multirow{8}{*}{ComCO}    
    &  \textit{ViT-H-14 (DFN)}& 67.08 & 22.19 & 10.73 & - & - \\
    & & \textit{ViT-SO400M-SigLIP} & 61.11 & 23.33 & 15.56 & - & - \\
    & & \textit{ViT-L-14 (datacomp)}& 72.23 & 19.05 & 8.72 & - & - \\
    & & \textit{xlm-roberta-large-ViT-H-14}& 43.60 & 31.36 & 25.05 & - & - \\
    & & \textit{ViT-L-14 (laion2b)} & 66.85 & 23.52 & 9.63 & - & - \\
    & & \textit{ViT-L-14 (openai)} & 57.66 & 26.75 & 15.59 & - & - \\
    & & \textit{ViT-B-32 (openai)} & 55.73 & 28.28 & 15.98 & - & - \\
    & & \textit{NegCLIP} & 57.56 & 29.45 & 12.99 & - & - \\
     
    \midrule
    \multirow{16}{*}{n = 4} & \multirow{8}{*}{SimCO}   
    
    &  \textit{ViT-H-14 (DFN)}& 60.06 & 12.77 & 12.03 & 15.14 & - \\
    & & \textit{ViT-SO400M-SigLIP} & 53.54 & 14.76 & 11.43 & 20.27 & - \\
    & & \textit{ViT-L-14 (datacomp)}& 62.16 & 15.99 & 10.41 & 11.44 & - \\
    & & \textit{xlm-roberta-large-ViT-H-14}& 62.58 & 22.52 & 10.91 & 3.99 & - \\
    & & \textit{ViT-L-14 (laion2b)} & 67.81 & 8.97 & 5.80 & 17.41 & - \\
    & & \textit{ViT-L-14 (openai)} & 66.87 & 11.59 & 6.18 & 15.35 & - \\
    & & \textit{ViT-B-32 (openai)} & 76.37 & 10.03 & 7.50 & 6.55 & - \\
    & & \textit{NegCLIP} & 82.90 & 10.20 & 4.61 & 2.29 & - \\

    \cmidrule{2-8}
    & \multirow{8}{*}{ComCO}    
    &  \textit{ViT-H-14 (DFN)}& 64.34 & 19.25 & 11.14 & 5.27 & - \\
    & & \textit{ViT-SO400M-SigLIP} & 58.11 & 21.16 & 10.99 & 9.73 & - \\
    & & \textit{ViT-L-14 (datacomp)}& 71.13 & 16.26 & 8.74 & 3.87 & - \\
    & & \textit{xlm-roberta-large-ViT-H-14}& 34.03 & 28.73 & 21.07 & 16.18 & - \\
    & & \textit{ViT-L-14 (laion2b)} & 63.96 & 21.59 & 10.68 & 3.76 & - \\
    & & \textit{ViT-L-14 (openai)} & 48.20 & 26.01 & 10.74 & 8.74 & - \\
    & & \textit{ViT-B-32 (openai)} & 50.31 & 20.74 & 15.45 & 6.79 & - \\
    & & \textit{NegCLIP} & 51.63 & 28.92 & 14.86 & 4.59 & - \\
     
    \midrule
    \multirow{16}{*}{n = 5}& \multirow{8}{*}{SimCO}

    &  \textit{ViT-H-14 (DFN)}& 60.80 & 10.61 & 8.35 & 9.02 & 11.22 \\
    & & \textit{ViT-SO400M-SigLIP} & 49.47 & 13.32 & 3.39 & 11.97 & 21.25 \\
    & & \textit{ViT-L-14 (datacomp)}& 66.43 & 16.12 & 6.59 & 4.99 & 5.87 \\
    & & \textit{xlm-roberta-large-ViT-H-14}& 60.65 & 21.03 & 11.90 & 5.15 & 1.28 \\
    & & \textit{ViT-L-14 (laion2b)} & 74.07 & 9.51 & 4.48 & 2.80 & 9.14 \\
    & & \textit{ViT-L-14 (openai)} & 71.71 & 10.59 & 2.99 & 2.71 & 12.00 \\
    & & \textit{ViT-B-32 (openai)} & 43.86 & 26.41 & 15.44 & 8.57 & 5.72 \\
    & & \textit{NegCLIP} & 85.00 & 10.39 & 3.12 &1.24 & 0.26 \\

    \cmidrule{2-8}
    & \multirow{8}{*}{ComCO}    
    &  \textit{ViT-H-14 (DFN)}& 61.06 & 17.00 & 11.98 & 6.69 & 3.27 \\
    & & \textit{ViT-SO400M-SigLIP} & 55.77 & 19.25 & 10.24 & 6.73 & 8.01 \\
    & & \textit{ViT-L-14 (datacomp)}& 68.96 & 14.61 & 9.40 & 4.77 & 2.25 \\
    & & \textit{xlm-roberta-large-ViT-H-14}& 28.86 & 26.87 & 19.42 & 14.61 & 10.24 \\
    & & \textit{ViT-L-14 (laion2b)} & 61.93 & 19.10 & 11.65 & 5.11 & 2.21 \\
    & & \textit{ViT-L-14 (openai)} & 38.40 & 24.80 & 18.79 & 11.04 & 6.68 \\
    & & \textit{ViT-B-32 (openai)} & 44.71 & 26.69 & 16.44 & 8.37 & 3.79 \\
    & & \textit{NegCLIP} & 45.70 & 27.56 & 17.03 & 7.57 & 2.15 \\

    \bottomrule
    \end{tabular}
    \end{table*}

\section{Image-based Object Classification}
\label{app:ioc}
\begin{table*}[htbp]
    \centering
    \scriptsize
    \setlength{\tabcolsep}{4pt}
    \renewcommand{\arraystretch}{1.1}
    \caption{Image-based Object Classification}
    \label{tab:ioc_total}
    \begin{tabular}{lllccccc}
    \toprule
    \rowcolor[HTML]{E4E8F2}
    Number of Objects & Dataset & Model & \textbf{Large Object} & \textbf{Small Obj 1} & \textbf{Small Obj 2} & \textbf{Small Obj 3} & \textbf{Small Obj 4} \\ 
    \midrule
    \multirow{16}{*}{n = 2}  & \multirow{8}{*}{SimCO}    
    &  \textit{ViT-H-14 (DFN)}& 88.1 & 14.29 & - & - & - \\
    & & \textit{ViT-SO400M-SigLIP} & 97.62 & 16.67 & - & - & - \\
    & & \textit{ViT-L-14 (datacomp)}& 83.33 & 11.9 & - & - & -  \\
    & & \textit{xlm-roberta-large-ViT-H-14}& 78.57 & 21.43 & - & - & -  \\
    & & \textit{ViT-L-14 (laion2b)} & 66.67 & 11.9 & - & - & -  \\
    & & \textit{ViT-L-14 (openai)} & 64.29 & 0.00 & - & - & -  \\
    & & \textit{ViT-B-32 (openai)} & 61.9 & 0.00 & - & - & -  \\
    & & \textit{NegCLIP} & 40.48 & 7.14 & - & - & - \\

    \cmidrule{2-8}
    
    & \multirow{8}{*}{ComCO}    
    &  \textit{ViT-H-14 (DFN)}& 100.0 & 26.36 & - & - & - \\
    & & \textit{ViT-SO400M-SigLIP} & 100.0 & 33.9 & - & - & - \\
    & & \textit{ViT-L-14 (datacomp)}& 100.0 & 42.35 & - & - & - \\
    & & \textit{xlm-roberta-large-ViT-H-14}& 100.0 & 40.85 & - & - & - \\
    & & \textit{ViT-L-14 (laion2b)} & 100.0 & 31.29 & - & - & - \\
    & & \textit{ViT-L-14 (openai)} & 99.8 & 41.29 & - & - & - \\
    & & \textit{ViT-B-32 (openai)} & 99.8 & 35.81 & - & - & - \\
    & & \textit{NegCLIP} & 99.6 & 41.95 & - & - & - \\
    
    \midrule 
    \multirow{16}{*}{n = 3} & \multirow{8}{*}{SimCO}   
    &  \textit{ViT-H-14 (DFN)}& 100.0 & 35.65 & 41.57 & - & -  \\
    & & \textit{ViT-SO400M-SigLIP} & 99.8 & 42.8 & 49.03 & - & -  \\
    & & \textit{ViT-L-14 (datacomp)}& 100.0 & 39.94 & 51.28 & - & -  \\
    & & \textit{xlm-roberta-large-ViT-H-14}& 99.9 & 48.42 & 56.28 & - & -  \\
    & & \textit{ViT-L-14 (laion2b)} & 99.8 & 45.56 & 56.08 & - & -  \\
    & & \textit{ViT-L-14 (openai)} & 98.98 & 39.73 & 50.46 & - & -  \\
    & & \textit{ViT-B-32 (openai)} & 96.12 & 38.1 & 51.58 & - & -  \\
    & & \textit{NegCLIP} & 97.04 & 42.59 & 59.35 & - & -  \\

    \cmidrule{2-8}
    & \multirow{8}{*}{ComCO}    
    &  \textit{ViT-H-14 (DFN)}& 100.0 & 29.12 & 21.5 & - & - \\
    & & \textit{ViT-SO400M-SigLIP} & 100.0 & 30.94 & 29.94 & - & - \\
    & & \textit{ViT-L-14 (datacomp)}& 100.0 & 36.56 & 33.5 & - & - \\
    & & \textit{xlm-roberta-large-ViT-H-14}& 100.0 & 33.69 & 32.31 & - & - \\
    & & \textit{ViT-L-14 (laion2b)} & 100.0 & 35.44 & 30.31 & - & - \\
    & & \textit{ViT-L-14 (openai)} & 99.94 & 33.31 & 34.31 & - & - \\
    & & \textit{ViT-B-32 (openai)} & 99.94 & 29.0 & 32.94 & - & - \\
    & & \textit{NegCLIP} & 99.81 & 33.88 & 43.0 & - & - \\
     
    \midrule
    \multirow{16}{*}{n = 4} & \multirow{8}{*}{SimCO}   
    
    &  \textit{ViT-H-14 (DFN)}& 100.0 & 40.06 & 34.06 & 41.31 & -  \\
    & & \textit{ViT-SO400M-SigLIP} & 100.0 & 47.0 & 38.5 & 41.06 & -  \\
    & & \textit{ViT-L-14 (datacomp)}& 100.0 & 48.94 & 38.38 & 45.06 & - \\
    & & \textit{xlm-roberta-large-ViT-H-14}& 100.0 & 48.19 & 35.81 & 46.38 & -  \\
    & & \textit{ViT-L-14 (laion2b)} & 100.0 & 50.5 & 41.81 & 43.94 & - \\
    & & \textit{ViT-L-14 (openai)} & 100.0 & 45.19 & 38.38 & 39.0  & - \\
    & & \textit{ViT-B-32 (openai)} & 100.0 & 38.06 & 31.5 & 37.25  & - \\
    & & \textit{NegCLIP} & 100.0 & 42.0 & 37.25 & 46.94  & - \\

    \cmidrule{2-8}
    & \multirow{8}{*}{ComCO}    
    &  \textit{ViT-H-14 (DFN)}& 100.0 & 16.64 & 14.13 & 12.38 & - \\
    & & \textit{ViT-SO400M-SigLIP} & 100.0 & 18.95 & 15.57 & 17.57 & - \\
    & & \textit{ViT-L-14 (datacomp)}& 100.0 & 20.64 & 21.01 & 19.01 & - \\
    & & \textit{xlm-roberta-large-ViT-H-14}& 100.0 & 20.45 & 18.45 & 16.51 & - \\
    & & \textit{ViT-L-14 (laion2b)} & 100.0 & 19.76 & 17.57 & 18.89 & - \\
    & & \textit{ViT-L-14 (openai)} & 99.94 & 19.32 & 21.89 & 22.39 & - \\
    & & \textit{ViT-B-32 (openai)} & 100.0 & 21.58 & 21.83 & 22.26 & - \\
    & & \textit{NegCLIP} & 100.0 & 21.89 & 23.64 & 31.33 & - \\
     
    \midrule
    \multirow{16}{*}{n = 5}& \multirow{8}{*}{SimCO}

    &  \textit{ViT-H-14 (DFN)}& 100.0 & 34.0 & 30.0 & 30.38 & 21.62 \\
    & & \textit{ViT-SO400M-SigLIP} & 100.0 & 38.5 & 34.7 & 27.38 & 25.62 \\
    & & \textit{ViT-L-14 (datacomp)}& 100.0 & 40.38 & 36.12 & 32.0 & 24.75 \\
    & & \textit{xlm-roberta-large-ViT-H-14}& 100.0 & 41.56 & 39.56 & 36.69 & 32.81 \\
    & & \textit{ViT-L-14 (laion2b)} & 100.0 & 43.88 & 39.5 & 34.0 & 28.94 \\
    & & \textit{ViT-L-14 (openai)} & 100.0 & 42.19 & 36.38 & 32.81 & 31.94 \\
    & & \textit{ViT-B-32 (openai)} & 98.81 & 36.25 & 35.38 & 33.88 & 26.06 \\
    & & \textit{NegCLIP} & 99.19 & 40.88 & 37.94 & 37.56 & 28.94 \\

    \cmidrule{2-8}
    & \multirow{8}{*}{ComCO}    
    &  \textit{ViT-H-14 (DFN)}& 100.0 & 13.88 & 9.38 & 9.32 & 11.94 \\
    & & \textit{ViT-SO400M-SigLIP} & 100.0 & 15.51 & 13.88 & 14.57 & 14.76 \\
    & & \textit{ViT-L-14 (datacomp)}& 100.0 & 18.2 & 15.07 & 16.07 & 18.32 \\
    & & \textit{xlm-roberta-large-ViT-H-14}& 99.94 & 15.38 & 14.88 & 15.26 & 19.14 \\
    & & \textit{ViT-L-14 (laion2b)} & 100.0 & 15.51 & 12.32 & 14.13 & 17.95 \\
    & & \textit{ViT-L-14 (openai)} & 100.0 & 15.38 & 14.76 & 16.76 & 20.01 \\
    & & \textit{ViT-B-32 (openai)} & 99.87 & 17.76 & 18.64 & 19.2 & 23.14 \\
    & & \textit{NegCLIP} & 100 & 18.89 & 16.57 & 23.51 & 28.77 \\

    \bottomrule
    \end{tabular}
\end{table*}

We conducted the IOC experiment on images from both generated datasets, focusing on scenarios where one object was significantly larger than the others. This experiment was repeated across various models. In our trials, we ensured that the larger object was not consistently placed in a fixed location, instead testing multiple positions. The average results of these experiments are presented in Table \ref{tab:ioc_total}.

\section{Image-based Object Retrieval}
\label{app:ior}
We extended our investigation by conducting the IOR experiment on images from both the SimCO and ComCO datasets. This experiment encompassed all scenarios ranging from 2 to 5 objects. Similar to our previous experiment, we deliberately varied the position of the larger object to avoid location-based biases. By considering different locations for the larger object, we aimed to better understand the impact of object size on the models' performance.
The results of these experiments are presented in Table \ref{tab:ior_total}.
\begin{table*}[htbp]
    \centering
    \scriptsize
    \setlength{\tabcolsep}{3pt}
    \renewcommand{\arraystretch}{1.1}
    \caption{Image-based Object Retrieval}
    \label{tab:ior_total}
    \begin{tabular}{llcccccc}
    \toprule
    \rowcolor[HTML]{E4E8F2}
    Number of Objects & Dataset & Model & \textbf{Large Object} & \textbf{Small Obj 1} & \textbf{Small Obj 2} & \textbf{Small Obj 3} & \textbf{Small Obj 4} \\ 
    \midrule
    \multirow{16}{*}{n = 2}  & \multirow{8}{*}{SimCO}    
    &  \textit{ViT-H-14 (DFN)}& 99.11 & 0.89 & - & - & - \\
    & & \textit{ViT-SO400M-SigLIP}& 91.67 & 8.33 & - & - & -  \\
    & & \textit{ViT-L-14 (datacomp)}& 91.96 & 8.04 & - & - & - \\
    & & \textit{xlm-roberta-large-ViT-H-14}& 94.92 & 5.08 & - & - & - \\
    & & \textit{ViT-L-14 (laion2b)}& 92.86 & 7.14 & - & - & - \\
    & & \textit{ViT-L-14 (openai)}& 87.88 & 12.12 & - & - & - \\
    & & \textit{ViT-B-32 (openai)}& 90.24 & 9.76 & - & - & - \\
    & & \textit{NegCLIP}& 94.64 & 5.36 & - & - & - \\

    \cmidrule{2-8}
    
    & \multirow{8}{*}{ComCO}    
    &  \textit{ViT-H-14 (DFN)}& 97.35 & 2.65 & - & - & - \\
    & & \textit{ViT-SO400M-SigLIP}& 95.13 & 4.87 & - & - & - \\
    & & \textit{ViT-L-14 (datacomp)} & 89.85 & 10.15 & - & - & - \\
    & & \textit{xlm-roberta-large-ViT-H-14}& 93.89 & 6.11 & - & - & - \\
    & & \textit{ViT-L-14 (laion2b)}& 94.84 & 5.16 & - & - & - \\
    & & \textit{ViT-L-14 (openai)} & 83.7 & 16.30 & - & - & -\\
    & & \textit{ViT-B-32 (openai)}& 86.86 & 13.14 & - & - & - \\
    & & \textit{NegCLIP}& 83.3 & 16.7 & - & - & - \\
    
    \midrule 
    \multirow{16}{*}{n = 3} & \multirow{8}{*}{SimCO}   
    &  \textit{ViT-H-14 (DFN)}& 93.80 & 0.65 & 5.55 & - & - \\
    & & \textit{ViT-SO400M-SigLIP}& 83.27 & 5.61 & 11.12 & - & - \\
    & & \textit{ViT-L-14 (datacomp)}& 77.16 & 5.81 & 17.04 & - & - \\
    & & \textit{xlm-roberta-large-ViT-H-14} & 80.21 & 5.12 & 14.66 & - & -\\
    & & \textit{ViT-L-14 (laion2b)}& 76.57 & 9.57 & 13.86 & - & - \\
    & & \textit{ViT-L-14 (openai)}& 72.07 & 8.66 & 19.27 & - & - \\
    & & \textit{ViT-B-32 (openai)}& 61.14 & 14.69 & 24.17 & - & - \\
    & & \textit{NegCLIP}& 59.13 & 14.91 & 25.96 & - & - \\

    \cmidrule{2-8}
    & \multirow{8}{*}{ComCO}    
    &  \textit{ViT-H-14 (DFN)}& 96.52 & 1.71 & 17.8 & - & - \\
    & & \textit{ViT-SO400M-SigLIP}& 90.5 & 5.47 & 4.03 & - & - \\
    & & \textit{ViT-L-14 (datacomp)}& 89.65 & 6.09 & 4.26 & - & - \\
    & & \textit{xlm-roberta-large-ViT-H-14}& 91.39 & 4.92 & 3.69 & - & - \\
    & & \textit{ViT-L-14 (laion2b)}& 91.26 & 3.28 & 5.46 & - & - \\
    & & \textit{ViT-L-14 (openai)}& 74.2 & 12.79 & 13.01 & - & - \\
    & & \textit{ViT-B-32 (openai)}& 80.6 & 5.22 & 14.18 & - & - \\
    & & \textit{NegCLIP}& 76.36 & 10.47 & 13.18 & - & - \\
     
    \midrule
    \multirow{16}{*}{n = 4} & \multirow{8}{*}{SimCO}   
    
    &  \textit{ViT-H-14 (DFN)}& 99.5 & 0.0 & 0.0 & 0.5 & - \\
    & & \textit{ViT-SO400M-SigLIP}& 91.03 & 1.28 & 2.99 & 4.7 & - \\
    & & \textit{ViT-L-14 (datacomp)} & 89.71 & 3.43 & 3.61 & 3.25 & -\\
    & & \textit{xlm-roberta-large-ViT-H-14}& 92.47 & 2.08 & 2.60 & 2.86 & - \\
    & & \textit{ViT-L-14 (laion2b)} & 86.92 & 4.67 & 3.74 & 4.67 & -\\
    & & \textit{ViT-L-14 (openai)}& 70.55 & 13.01 & 7.53 & 8.9 & - \\
    & & \textit{ViT-B-32 (openai)}& 52.17 & 18.84 & 13.04 & 15.94 & - \\
    & & \textit{NegCLIP}& 74.4 & 10.4 & 7.2 & 8.0 & - \\

    \cmidrule{2-8}
    & \multirow{8}{*}{ComCO}    
    &  \textit{ViT-H-14 (DFN)}& 95.86 & 2.55 & 1.27 & 0.32 & - \\
    & & \textit{ViT-SO400M-SigLIP}& 94.03 & 2.24 & 1.49 & 2.24 & - \\
    & & \textit{ViT-L-14 (datacomp)}& 93.3 & 3.91 & 1.12 & 16.8 & - \\
    & & \textit{xlm-roberta-large-ViT-H-14} & 90.91 & 2.02 & 5.05 & 2.02 & -\\
    & & \textit{ViT-L-14 (laion2b)} & 91.78 & 5.48 & 2.74 & 0.0 & -\\
    & & \textit{ViT-L-14 (openai)}& 67.86 & 14.29 & 7.14 & 10.71 & - \\
    & & \textit{ViT-B-32 (openai)}& 85.0 & 0.0 & 5.0 & 10.0 & - \\
    & & \textit{NegCLIP}& 79.55 & 0.0 & 2.27 & 18.19 & - \\
     
    \midrule
    \multirow{16}{*}{n = 5}& \multirow{8}{*}{SimCO}

    &  \textit{ViT-H-14 (DFN)}& 100.0 & 0.0 & 0.0 & 0.0 & 0.0 \\
    & & \textit{ViT-SO400M-SigLIP}& 94.92 & 3.39 & 1.69 & 0.0 & 0.0 \\
    & & \textit{ViT-L-14 (datacomp)}& 91.3 & 5.59 & 1.24 & 1.24 & 0.62 \\
    & & \textit{xlm-roberta-large-ViT-H-14}& 77.42 & 11.83 & 5.38 & 3.23 & 2.15 \\
    & & \textit{ViT-L-14 (laion2b)}& 81.01 & 8.86 & 5.06 & 1.27 & 0.38 \\
    & & \textit{ViT-L-14 (openai)}& 77.14 & 8.57 & 5.71 & 5.71 & 2.86 \\
    & & \textit{ViT-B-32 (openai)}& 68.75 & 25.0 & 6.25 & 0.0 & 0.0 \\
    & & \textit{NegCLIP}& 58.62 & 17.24 & 15.52 & 5.17 & 3.45 \\

    \cmidrule{2-8}
    & \multirow{8}{*}{ComCO}    
    &  \textit{ViT-H-14 (DFN)}& 95.16 & 1.61  & 1.61 & 0.0 & 1.61 \\
    & & \textit{ViT-SO400M-SigLIP}& 80.0 & 0.0 & 0.0 & 0.0 & 20.0 \\
    & & \textit{ViT-L-14 (datacomp)}& 90.91 & 4.55 & 0.0 & 0.0 & 4.55 \\
    & & \textit{xlm-roberta-large-ViT-H-14}& 100.0 & 0.0 & 0.0 & 0.0 & 0.0 \\
    & & \textit{ViT-L-14 (laion2b)}& 100.0 & 0.0 & 0.0 & 0.0 & 0.0 \\
    & & \textit{ViT-L-14 (openai)} & 100.0 & 0.0 & 0.0 & 0.0 & 0.0 \\
    & & \textit{ViT-B-32 (openai)}& 100.0 & 0.0 & 0.0 & 0.0 & 0. \\
    & & \textit{NegCLIP}& 50.0 & 0.0 & 0.0 & 50.0 & 0.0 \\

    \bottomrule
    \end{tabular}
\end{table*}

\section{Text-based Object Classification for Long Caption}
\label{app:toc-long}

In this section, we revisited the IOC experiment with a significant modification to the caption structure. Our objective was to investigate whether the previously observed bias persists in longer, more elaborate captions. We achieved this by expanding the caption template, incorporating additional descriptive phrases between object mentions.

The extended caption template used in this experiment was as follows:

\begin{quote}
This vibrant display features a stunning OBJ1 with its radiant glow, a mesmerizing OBJ2 with bold contours, an enchanting OBJ3 that fits perfectly with its graceful form, a dazzling OBJ4 with brilliant tones and intricate patterns, and an alluring OBJ5 that completes the ensemble with its seamless fusion and distinct shape.
\end{quote}

This template allowed us to maintain a consistent structure while significantly increasing the caption length and complexity.

The results of this modified IOC experiment are presented in Table \ref{tab:toc_long_total}. Notably, the observed pattern closely resembles that of the standard IOC experiment. This similarity suggests that the bias identified in shorter captions persists even in more elaborate textual descriptions.

\begin{table*}[ht]
    \centering
    \scriptsize
    \setlength{\tabcolsep}{4pt}
    \renewcommand{\arraystretch}{1.1}
    \caption{Text-based Object Classification on Long Captions}
    \label{tab:toc_long_total}
    \begin{tabular}{lllccccc}
    \toprule
    \rowcolor[HTML]{E4E8F2}
    Number of Objects & Dataset & Model & \textbf{First Object} & \textbf{Second Object} & \textbf{Third Object} & \textbf{Fourth Object} & \textbf{Fifth Object} \\ 
    \midrule
    \multirow{16}{*}{n = 2}  & \multirow{8}{*}{SimCO}    
    &  \textit{ViT-H-14 (DFN)}& 100.0 & 89.01 & - & - & - \\
    & & \textit{ViT-SO400M-SigLIP} & 100.0 & 93.83 & - & - & - \\
    & & \textit{ViT-L-14 (datacomp)}& 100.0 & 63.22 & - & - & - \\
    & & \textit{xlm-roberta-large-ViT-H-14}& 99.82 & 51.83 & - & - & - \\
    & & \textit{ViT-L-14 (laion2b)} & 100.0 & 85.88 & - & - & - \\
    & & \textit{ViT-L-14 (openai)} & 99.65 & 98.26 & - & - & - \\
    & & \textit{ViT-B-32 (openai)} & 100.0 & 72.69 & - & - & - \\
    & & \textit{NegCLIP} & 100 & 89.59  & - & - & - \\

    \cmidrule{2-8}
    
    & \multirow{8}{*}{ComCO}    
    &  \textit{ViT-H-14 (DFN)}& 99.99 & 99.86 & - & - & - \\
    & & \textit{ViT-SO400M-SigLIP} & 100 & 99.48 & - & - & - \\
    & & \textit{ViT-L-14 (datacomp)}& 100 & 98.89 & - & - & - \\
    & & \textit{xlm-roberta-large-ViT-H-14}& 99.95 & 92.84 & - & - & - \\
    & & \textit{ViT-L-14 (laion2b)} & 100 & 99.03 & - & - & - \\
    & & \textit{ViT-L-14 (openai)} & 99.99 & 99.99 & - & - & - \\
    & & \textit{ViT-B-32 (openai)} & 99.59 & 99.45 & - & - & - \\
    & & \textit{NegCLIP} & 99.94 & 98.99 & - & - & - \\
    
    \midrule 
    \multirow{16}{*}{n = 3} & \multirow{8}{*}{SimCO}   
    &  \textit{ViT-H-14 (DFN)}& 99.34 & 43.49 & 89.66 & - & - \\
    & & \textit{ViT-SO400M-SigLIP} & 100.0 & 65.26 & 49.76 & - & - \\
    & & \textit{ViT-L-14 (datacomp)}& 100.0 & 30.47 & 37.20 & - & - \\
    & & \textit{xlm-roberta-large-ViT-H-14}& 97.78 & 22.96 & 27.23 & - & - \\
    & & \textit{ViT-L-14 (laion2b)} & 99.65 & 57.67 & 35.51 & - & - \\
    & & \textit{ViT-L-14 (openai)} & 99.13 & 86.67  & 58.22 & - & - \\
    & & \textit{ViT-B-32 (openai)} & 96.26 & 54.19 & 44.88 & - & - \\
    & & \textit{NegCLIP} & 98.30 & 67.60 & 65.90 & - & - \\

    \cmidrule{2-8}
    & \multirow{8}{*}{ComCO}    
    &  \textit{ViT-H-14 (DFN)}& 99.31 & 78.44 & 84.15 & - & - \\
    & & \textit{ViT-SO400M-SigLIP} & 99.93 & 67.22 & 76.89 & - & - \\
    & & \textit{ViT-L-14 (datacomp)}& 98.98 & 85.77 & 65.64 & - & - \\
    & & \textit{xlm-roberta-large-ViT-H-14}& 99.21 & 38.60 & 60.10 & - & - \\
    & & \textit{ViT-L-14 (laion2b)} & 98.81 & 82.72 & 74.31 & - & - \\
    & & \textit{ViT-L-14 (openai)} & 99.41 & 96.44 & 82.18 & - & - \\
    & & \textit{ViT-B-32 (openai)} & 95.59 & 81.91 & 76.09 & - & - \\
    & & \textit{NegCLIP} & 98.62 & 74.29 & 81.70 & - & - \\
     
    \midrule
    \multirow{16}{*}{n = 4} & \multirow{8}{*}{SimCO}   
    
    &  \textit{ViT-H-14 (DFN)}& 99.17 & 24.74 & 67.00 & 41.46 & - \\
    & & \textit{ViT-SO400M-SigLIP} & 100.0 & 46.75 & 24.40 & 20.93 & - \\
    & & \textit{ViT-L-14 (datacomp)}& 100.0 & 15.27 & 17.79 & 43.03 & - \\
    & & \textit{xlm-roberta-large-ViT-H-14}& 98.87 & 13.34 & 12.67 & 15.85 & - \\
    & & \textit{ViT-L-14 (laion2b)} & 99.56 & 36.03 & 19.23 & 34.51 & - \\
    & & \textit{ViT-L-14 (openai)} & 98.22 & 70.29 & 40.54 & 50.71 & - \\
    & & \textit{ViT-B-32 (openai)} & 97.47 & 41.20 & 25.18 & 24.31 & - \\
    & & \textit{NegCLIP} & 98.93 & 49.58 & 35.89 & 35.40 & - \\

    \cmidrule{2-8}
    & \multirow{8}{*}{ComCO}    
    &  \textit{ViT-H-14 (DFN)}& 98.34 & 62.49 & 70.25 & 42.34 & - \\
    & & \textit{ViT-SO400M-SigLIP} & 99.90 & 39.28 & 58.01 & 32.51 & - \\
    & & \textit{ViT-L-14 (datacomp)}& 97.95 & 71.61 & 37.24 & 48.50 & - \\
    & & \textit{xlm-roberta-large-ViT-H-14} & 99.34 & 20.38 & 21.45 & 25.08 & - \\
    & & \textit{ViT-L-14 (laion2b)}& 98.41 & 66.90 & 51.43 & 38.87 & - \\
    & & \textit{ViT-L-14 (openai)} & 96.39 & 88.74 & 62.87 & 75.1 & - \\
    & & \textit{ViT-B-32 (openai)} & 96.81 & 62.50 & 59.19 & 22.93 & - \\
    & & \textit{NegCLIP} & 98.50 & 45.93 & 40.11 & 68.58 & - \\
     
    \midrule
    \multirow{16}{*}{n = 5}& \multirow{8}{*}{SimCO}

    &  \textit{ViT-H-14 (DFN)}& 97.44 & 18.82 & 53.68  & 26.08 & 47.45\\
    & & \textit{ViT-SO400M-SigLIP} & 100.0 & 20.35 & 19.30 & 12.57 & 18.40 \\
    & & \textit{ViT-L-14 (datacomp)}& 99.74 & 17.57 & 19.29 & 41.34 & 23.67 \\
    & & \textit{xlm-roberta-large-ViT-H-14}& 99.09 & 12.51 & 8.49 & 8.63 & 30.25 \\
    & & \textit{ViT-L-14 (laion2b)} &  99.69 & 60.13 & 28.18  & 49.20 & 54.92\\
    & & \textit{ViT-L-14 (openai)} & 96.26 & 70.36 & 44.68 & 36.7 & 48.1 \\
    & & \textit{ViT-B-32 (openai)} & 96.79 & 30.71 & 15.25 & 12.58 & 41.30 \\
    & & \textit{NegCLIP} & 99.35 & 32.26 & 22.22 & 16.39 & 62.63 \\

    \cmidrule{2-8}
    & \multirow{8}{*}{ComCO}    
    &  \textit{ViT-H-14 (DFN)}& 97.45 & 43.49 & 29.20 & 17.91 & 1.13 \\
    & & \textit{ViT-SO400M-SigLIP} & 98.46 & 45.21 & 32.54 & 26.64 & 1.18 \\
    & & \textit{ViT-L-14 (datacomp)}& 92.76 & 40.83 & 17.56 & 9.8 & 1.05 \\
    & & \textit{xlm-roberta-large-ViT-H-14}& 99.84 & 13.18 & 11.02 & 8.26 & 45.38 \\
    & & \textit{ViT-L-14 (laion2b)} & 97.39 & 41.48 & 19.5 & 9.4 & 1.26 \\
    & & \textit{ViT-L-14 (openai)} & 92.81 & 68.46 & 31.85 & 9.8 & 1.24 \\
    & & \textit{ViT-B-32 (openai)} & 95.85 & 42.62 & 22.24 & 9.18 & 0.9 \\
    & & \textit{NegCLIP} & 99.16 & 27.60 & 19.78 & 21.80 & 69.08 \\

    \bottomrule
    \end{tabular}
\end{table*}

\section{Text-based Object Retrieval for Long Caption}
\label{app:tor-long}
\begin{table*}[htbp]
    \centering
    \scriptsize
    \setlength{\tabcolsep}{3pt}
    \renewcommand{\arraystretch}{1.1}
    \caption{Text-based Object Retrieval For long template}
    \label{tab:tor_long_total}
    \begin{tabular}{lllcccccc}
    \toprule
    \rowcolor[HTML]{E4E8F2}
    Number of Objects & Dataset & Model & \textbf{Accuracy} & \textbf{First Object} & \textbf{Second Object} & \textbf{Third Object} & \textbf{Fourth Object} & \textbf{Fifth Object} \\ 
    \midrule
    \multirow{16}{*}{n = 2}  & \multirow{8}{*}{SimCO}       
    &  \textit{ViT-H-14 (DFN)}& 96.73 & 62.16 & 37.84 & - & - & -\\
    & & \textit{ViT-SO400M-SigLIP} & 5.88 & 100.0 & 0.00 & - & - & -\\
    & & \textit{ViT-L-14 (datacomp)}& 98.04 & 70.67 & 29.33 & - & - & - \\
    & & \textit{xlm-roberta-large-ViT-H-14}& 98.69 & 76.82 & 23.18 & - & - & - \\
    & & \textit{ViT-L-14 (laion2b)} & 51.63 & 62.03 & 37.97 & - & - & - \\
    & & \textit{ViT-L-14 (openai)} & 96.08 & 39.46 & 60.54 & - & - & - \\
    & & \textit{ViT-B-32 (openai)} & 79.74 & 45.90 & 54.10 & - & - & - \\
    & & \textit{NegCLIP} & 99.35 & 38.82 & 61.18 & - & - & -\\

    \cmidrule{2-9}
    
    & \multirow{8}{*}{ComCO}    
    &  \textit{ViT-H-14 (DFN)}& 92.38 & 71.03 & 28.97 & - & - & -\\
    & & \textit{ViT-SO400M-SigLIP} & 3.42 & 100.0 & 0.00 & - & - & -\\
    & & \textit{ViT-L-14 (datacomp)}& 84.32 & 62.63 & 37.37 & - & - & -\\
    & & \textit{xlm-roberta-large-ViT-H-14}& 72.06 & 63.31 & 36.69 & - & - & -\\
    & & \textit{ViT-L-14 (laion2b)} & 58.73 & 63.01 & 36.99 & - & - & -\\
    & & \textit{ViT-L-14 (openai)} & 84.64 & 61.27 & 38.70 & - & - & -\\
    & & \textit{ViT-B-32 (openai)} & 78.38 & 61.77 & 37.78 & - & - & -\\
    & & \textit{NegCLIP} & 82.67 & 55.63 & 44.37 & - & - & -\\
    
    \midrule 
    \multirow{16}{*}{n = 3} & \multirow{8}{*}{SimCO}   
    &  \textit{ViT-H-14 (DFN)}& 88.6 & 43.02 & 30.43 & 26.56 & -  & -\\
    & & \textit{ViT-SO400M-SigLIP} & 0.74 & 100.0 & 0.00 & 0.00 & -  & -\\
    & & \textit{ViT-L-14 (datacomp)}& 88.48 & 63.02 & 24.38 & 12.60 & -  & -\\
    & & \textit{xlm-roberta-large-ViT-H-14}& 89.83 & 61.66 & 22.10 & 16.23 & -  & -\\
    & & \textit{ViT-L-14 (laion2b)} & 31.86 & 56.54 & 26.15 & 17.31 & -  & -\\
    & & \textit{ViT-L-14 (openai)} & 69.73 & 24.08 & 39.89 & 36.03 & -  & -\\
    & & \textit{ViT-B-32 (openai)} & 38.24 & 25.96 & 39.10 & 34.94 & -  & -\\
    & & \textit{NegCLIP} & 72.30 & 23.39 & 52.71 & 23.90 & -  & -\\

    \cmidrule{2-9}
    & \multirow{8}{*}{ComCO}    
    &  \textit{ViT-H-14 (DFN)}& 76.75 & 50.43 & 22.45 & 27.12 & - & -\\
    & & \textit{ViT-SO400M-SigLIP} & 0.07 & 100.0 & 0.00 & 0.00 & - & -\\
    & & \textit{ViT-L-14 (datacomp)}& 56.14 & 47.80 & 34.17 & 18.03 & - & -\\
    & & \textit{xlm-roberta-large-ViT-H-14}& 36.78 & 48.46 & 28.75 & 22.79 & - & -\\
    & & \textit{ViT-L-14 (laion2b)} & 29.17 & 48.75 & 35.78 & 15.47 & - & -\\
    & & \textit{ViT-L-14 (openai)} & 52.38 & 43.44 & 37.00 & 19.53 & - & -\\
    & & \textit{ViT-B-32 (openai)} & 49.97 & 47.58 & 30.75 & 21.45 & - & -\\
    & & \textit{NegCLIP} & 50.80 & 38.67 & 38.16 & 23.17 & - & -\\
     
    \midrule
    \multirow{16}{*}{n = 4} & \multirow{8}{*}{SimCO}   
    &  \textit{ViT-H-14 (DFN)}& 66.47 & 39.82 & 21.88 & 24.34 & 13.96  & -\\
    & & \textit{ViT-SO400M-SigLIP} & 0.49 & 100.0 & 0.00 & 0.00 & 0.00   & -\\
    & & \textit{ViT-L-14 (datacomp)}& 74.58 & 61.74 & 22.17 & 10.96 & 5.13  & -\\
    & & \textit{xlm-roberta-large-ViT-H-14}& 65.95 & 53.96 & 21.36 & 19.33 & 5.35  & -\\
    & & \textit{ViT-L-14 (laion2b)} & 22.42 & 66.76 & 17.78 & 11.22 & 4.23  & -\\
    & & \textit{ViT-L-14 (openai)} & 58.73 & 16.30 & 32.78 & 26.49 & 24.37  & -\\
    & & \textit{ViT-B-32 (openai)} & 18.43 & 35.64 & 37.77 & 14.18 & 12.41  & -\\
    & & \textit{NegCLIP} & 50.78 & 26.25 & 49.94 & 16.73 & 7.08  & -\\

    \cmidrule{2-9}
    & \multirow{8}{*}{ComCO}    
    &  \textit{ViT-H-14 (DFN)}& 52.87 & 47.87 & 20.54 & 22.72 & 8.87 & -\\
    & & \textit{ViT-SO400M-SigLIP} & 0.01 & 100.0 & 0.00 & 0.00 & 0.00 & -\\
    & & \textit{ViT-L-14 (datacomp)}& 31.36 & 39.21 & 30.74 & 20.94 & 9.11 & -\\
    & & \textit{xlm-roberta-large-ViT-H-14}& 14.99 & 43.03 & 24.29 & 19.72 & 12.96 & -\\
    & & \textit{ViT-L-14 (laion2b)} & 10.19 & 42.66 & 34.16 & 17.09 & 6.09 & -\\
    & & \textit{ViT-L-14 (openai)} & 28.78 & 35.25 & 31.55 & 19.19 & 13.86 & -\\
    & & \textit{ViT-B-32 (openai)} & 21.62 & 43.69 & 24.57 & 16.78 & 14.59 & -\\
    & & \textit{NegCLIP} & 19.41 & 30.36 & 30.38 & 24.39 & 14.86 & -\\
     
    \midrule
    \multirow{16}{*}{n = 5}& \multirow{8}{*}{SimCO}   
    
    &  \textit{ViT-H-14 (DFN)}& 45.44 & 43.46 & 20.45 & 18.34 & 11.87 & 5.88 \\
    & & \textit{ViT-SO400M-SigLIP} & 0.16 & 100.0 & 0.00 & 0.00 & 0.00 & 0.00 \\
    & & \textit{ViT-L-14 (datacomp)}& 51.45 & 59.26 & 22.46 & 8.12 & 8.46 & 1.70 \\
    & & \textit{xlm-roberta-large-ViT-H-14}& 52.92 & 54.87 & 13.81 & 19.30 & 8.16 & 3.86 \\
    & & \textit{ViT-L-14 (laion2b)} & 12.34 & 75.40 & 10.31 & 8.42 & 4.26 & 1.61 \\
    & & \textit{ViT-L-14 (openai)} & 29.39 & 8.98 & 29.39 & 28.44 & 15.97 & 17.20 \\
    & & \textit{ViT-B-32 (openai)} & 6.69 & 32.11 & 38.57 & 12.22 & 8.55 & 8.55 \\
    & & \textit{NegCLIP} & 17.54 & 23.15 & 41.18 & 24.48 & 7.65 & 3.53 \\

    \cmidrule{2-9}
    & \multirow{8}{*}{ComCO}    
    &  \textit{ViT-H-14 (DFN)}& 23.56 & 36.07 & 19.21 & 22.65 & 11.90 & 10.17 \\
    & & \textit{ViT-SO400M-SigLIP} & 0.00 & 100.0 & 0.00 & 0.00 & 0.00 & 0.00 \\
    & & \textit{ViT-L-14 (datacomp)}& 12.49 & 32.55 & 27.84 & 23.76 & 12.73 & 3.11 \\
    & & \textit{xlm-roberta-large-ViT-H-14}& 9.26 & 40.26 & 21.35 & 18.16 & 11.99 & 8.23 \\
    & & \textit{ViT-L-14 (laion2b)} & 4.57 & 38.49 & 31.50 & 17.50 & 8.31 & 4.20 \\
    & & \textit{ViT-L-14 (openai)} & 1.75 & 21.59 & 18.57 & 20.25 & 20.54 & 19.02 \\
    & & \textit{ViT-B-32 (openai)} & 1.86 & 32.72 & 15.62 & 14.71 & 18.36 & 16.26 \\
    & & \textit{NegCLIP} & 1.41 & 24.30 & 23.17 & 22.14 & 17.64 & 12.75 \\

    \bottomrule
    \end{tabular}
    \end{table*}

In this section, we aimed to examine the performance of various models in the IOR experiment when presented with longer caption formats. This approach mirrors our previous investigation, allowing us to draw comparisons between standard and extended caption scenarios.

We utilized the same extended caption template as in the previous section.
The results of this experiment are presented in Table \ref{tab:tor_long_total}. Notably, the observed pattern closely aligns with that of the standard IOR experiment, suggesting a consistency in model behavior across different caption lengths.

\section{Image-text matching}
\label{app:imtexmatch}
In this section, we extended the experiment previously conducted in Section 5.1, broadening its scope to encompass both the SimCO and ComCO datasets. Our investigation covered scenarios involving 2 to 5 objects and was replicated across various models.
The results of this comprehensive experiment are presented in Table \ref{tab:imgtxtmatch_long_total}.
\begin{table*}[htbp]
    \centering
    \scriptsize
    \setlength{\tabcolsep}{4pt}
    \renewcommand{\arraystretch}{1.1}
    \caption{Image-Text Matching on SimCO and ComCO}
    \label{tab:imgtxtmatch_long_total}
    \begin{tabular}{lllcc}
    \toprule
    \rowcolor[HTML]{E4E8F2}
    Number of Objects & Dataset & Model & \textbf{Orginal} & \textbf{Reordered} \\ 
    \midrule
    \multirow{16}{*}{n = 2}  & \multirow{8}{*}{SimCO}    
    &  \textit{ViT-H-14 (DFN)}& 94.55 & 91.01 \\
    & & \textit{ViT-SO400M-SigLIP} & 91.70 & 88.81 \\
    & & \textit{ViT-L-14 (datacomp)}& 92.26 & 89.43 \\
    & & \textit{xlm-roberta-large-ViT-H-14}& 93.15 & 91.87 \\
    & & \textit{ViT-L-14 (laion2b)} & 88.93 & 87.23 \\
    & & \textit{ViT-L-14 (openai)} & 77.20 & 74.08 \\
    & & \textit{ViT-B-32 (openai)} & 68.15 & 64.40 \\
    & & \textit{NegCLIP} & 78.30 & 73.24  \\
    
    \cmidrule{2-5}
    
    & \multirow{8}{*}{ComCO}    
    &  \textit{ViT-H-14 (DFN)}& 94.27 & 89.44 \\
    & & \textit{ViT-SO400M-SigLIP} & 92.21 & 88.96 \\
    & & \textit{ViT-L-14 (datacomp)}& 92.11 & 88.22 \\
    & & \textit{xlm-roberta-large-ViT-H-14}& 91.87 & 87.48 \\
    & & \textit{ViT-L-14 (laion2b)} & 89.96 & 83.62 \\
    & & \textit{ViT-L-14 (openai)} & 87.32 & 83.56 \\
    & & \textit{ViT-B-32 (openai)} & 79.79 & 74.36 \\
    & & \textit{NegCLIP} & 81.32 & 72.62 \\
    
    \midrule 
    \multirow{16}{*}{n = 3} & \multirow{8}{*}{SimCO}   
    &  \textit{ViT-H-14 (DFN)}& 94.73 & 89.81 \\
    & & \textit{ViT-SO400M-SigLIP} & 93.69 & 89.10 \\
    & & \textit{ViT-L-14 (datacomp)}& 94.48 & 89.28 \\
    & & \textit{xlm-roberta-large-ViT-H-14}& 92.97 & 89.92 \\
    & & \textit{ViT-L-14 (laion2b)} & 90.47 & 85.29 \\
    & & \textit{ViT-L-14 (openai)} & 83.44 & 78.38 \\
    & & \textit{ViT-B-32 (openai)} & 78.09 & 73.24 \\
    & & \textit{NegCLIP} & 79.72 & 73.31 \\
    
    \cmidrule{2-5}
    
    & \multirow{8}{*}{ComCO}    
    &  \textit{ViT-H-14 (DFN)}& 91.66 & 80.22 \\
    & & \textit{ViT-SO400M-SigLIP} & 89.85 & 84.39 \\
    & & \textit{ViT-L-14 (datacomp)}& 89.64 & 80.26 \\
    & & \textit{xlm-roberta-large-ViT-H-14}& 87.99 & 79.73 \\
    & & \textit{ViT-L-14 (laion2b)} & 87.02 & 72.23 \\
    & & \textit{ViT-L-14 (openai)} & 83.96 & 74.93 \\
    & & \textit{ViT-B-32 (openai)} & 77.20 & 66.71 \\
    & & \textit{NegCLIP} & 79.30 & 62.65 \\
     
    \midrule
    \multirow{16}{*}{n = 4} & \multirow{8}{*}{SimCO}   
    &  \textit{ViT-H-14 (DFN)}& 88.36 & 74.70 \\
    & & \textit{ViT-SO400M-SigLIP} & 83.32 & 73.16 \\
    & & \textit{ViT-L-14 (datacomp)}& 87.81 & 73.62 \\
    & & \textit{xlm-roberta-large-ViT-H-14}& 86.64 & 77.65 \\
    & & \textit{ViT-L-14 (laion2b)} & 82.43 & 69.99 \\
    & & \textit{ViT-L-14 (openai)} & 74.04 & 63.42 \\
    & & \textit{ViT-B-32 (openai)} & 67.00 & 55.72 \\
    & & \textit{NegCLIP} & 69.01 & 55.06 \\
    
    \cmidrule{2-5}
    
    & \multirow{8}{*}{ComCO}    
    &  \textit{ViT-H-14 (DFN)}& 88.22 & 61.33 \\
    & & \textit{ViT-SO400M-SigLIP} & 84.73 & 74.09 \\
    & & \textit{ViT-L-14 (datacomp)}& 86.24 & 63.99 \\
    & & \textit{xlm-roberta-large-ViT-H-14}& 83.83 & 65.66 \\
    & & \textit{ViT-L-14 (laion2b)} & 82.82 & 50.62 \\
    & & \textit{ViT-L-14 (openai)} & 80.48 & 60.71 \\
    & & \textit{ViT-B-32 (openai)} & 73.22 & 52.23 \\
    & & \textit{NegCLIP} & 76.68 & 46.94 \\
     
    \midrule
    \multirow{16}{*}{n = 5}& \multirow{8}{*}{SimCO}   
    &  \textit{ViT-H-14 (DFN)}& 87.05 & 70.79 \\
    & & \textit{ViT-SO400M-SigLIP} & 82.91 & 67.62 \\
    & & \textit{ViT-L-14 (datacomp)}& 87.10 & 68.35 \\
    & & \textit{xlm-roberta-large-ViT-H-14}& 86.80 & 73.32 \\
    & & \textit{ViT-L-14 (laion2b)} & 82.35 & 64.79 \\
    & & \textit{ViT-L-14 (openai)} & 89.79 & 26.33 \\
    & & \textit{ViT-B-32 (openai)} & 71.39 & 56.50 \\
    & & \textit{NegCLIP} & 74.06 & 54.92 \\
    
    \cmidrule{2-5}
    
    & \multirow{8}{*}{ComCO}    
    &  \textit{ViT-H-14 (DFN)}& 85.95 & 46.99 \\
    & & \textit{ViT-SO400M-SigLIP} & 84.31 & 69.09 \\
    & & \textit{ViT-L-14 (datacomp)}& 99.13 & 14.75 \\
    & & \textit{xlm-roberta-large-ViT-H-14}& 84.74 & 51.65 \\
    & & \textit{ViT-L-14 (laion2b)} & 82.33 & 40.76 \\
    & & \textit{ViT-L-14 (openai)} & 78.12 & 50.94 \\
    & & \textit{ViT-B-32 (openai)} & 72.13 & 44.35 \\
    & & \textit{NegCLIP} & 75.33 & 37.50 \\
    
    \bottomrule
    \end{tabular}
\end{table*}

\section{COCO Dataset Analysis}
\label{app:coco-anlysis}
In this section, we repeated the experiment conducted in Section 4.3 for different scenarios involving 2 to 5 objects. We divided the captions in the COCO dataset into four subsets: those mentioning 2 objects, 3 objects, 4 objects, and 5 objects. We then analyzed each subset to determine in what percentage of cases the largest object appeared in which position.

\begin{figure*} [h!]
    \centering
    \includegraphics[width=\linewidth]{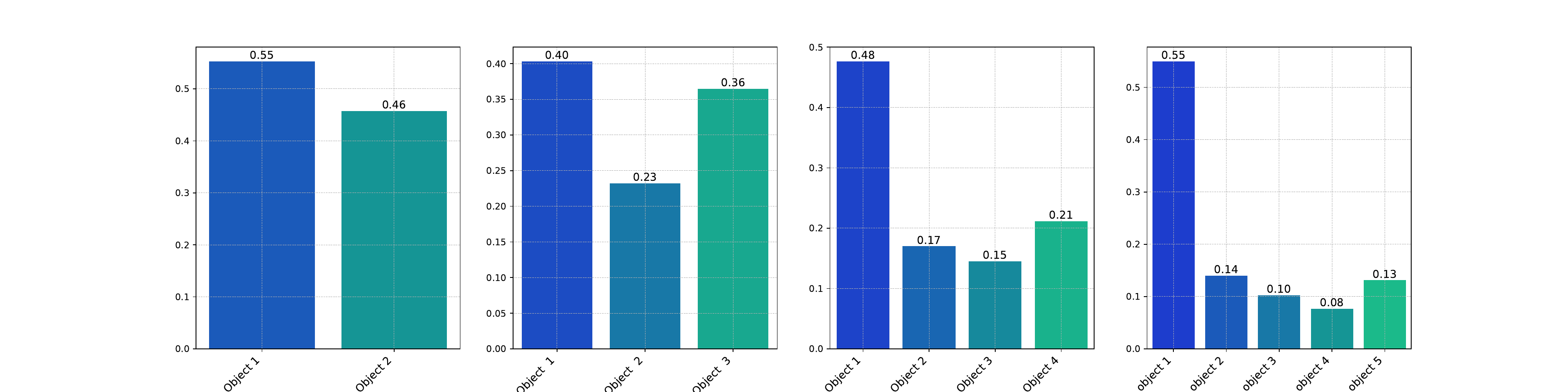}
    \caption{Distribution of larger object positions in captions for objects in COCO dataset}
    \label{fig:coco_analysis_total}
\end{figure*}

The results of this evaluation are presented in Figure \ref{fig:coco_analysis_total}. As can be observed, this trend is repeated across all scenarios: in most cases, the larger object appears earlier in the caption.

\end{document}